\documentclass[twocolumn]{article}
\usepackage{graphicx}
\usepackage{url}
\usepackage{amsmath}
\usepackage{amssymb}
\usepackage{amsfonts}
\usepackage{xcolor}
\usepackage{array}
\usepackage{colortbl}
\usepackage{float}
\usepackage{subcaption}
\usepackage{caption}
\usepackage{booktabs}
\usepackage{tabularx}
\usepackage{geometry}
\usepackage{multirow}
\usepackage{hyperref}
\usepackage{authblk}
\usepackage{algorithm}
\usepackage{algorithmic}

\title{Heterogeneous Variational Inference for Markov Degradation Hazard Models:
Discretized Mixture with Interpretable Clusters}

\author[1]{Takato Yasuno}
\affil[1]{Digital Business Department, Yachiyo Engineering Co., Ltd.}

\date{}

\begin{document}

\renewcommand{\abstractname}{Abstract}
\renewcommand{\refname}{References}
\renewcommand{\figurename}{Figure}
\renewcommand{\tablename}{Table}

\maketitle

\begin{abstract}

Infrastructure asset management requires understanding heterogeneous degradation patterns across equipment to optimize maintenance decisions. Traditional survival analysis assumes homogeneous hazard rates, failing to capture pump-specific degradation speeds that vary by 17$\times$ to 34$\times$ relative to the population average in real-world industrial settings. 

\textbf{Challenge:} Bayesian finite mixture models can identify discrete risk clusters (low-risk vs. high-risk equipment), but face three critical bottlenecks: (1) insufficient degradation signals from coarse state discretization, (2) unstable cluster identification when data inherently supports fewer clusters than explored, and (3) computational infeasibility of Markov Chain Monte Carlo (MCMC) methods for production deployment (7+ hours per model).

\textbf{Approach:} We propose a practical framework combining (1) \textbf{8-state global percentile discretization} that amplifies degradation events by +83\% (1.3\% $\rightarrow$ 2.4\% transition rate), (2) \textbf{30-dimensional feature engineering} integrating statistical trends (22 features), continuous health indicators, and text embeddings (PCA-compressed to 3 dimensions), (3) \textbf{interpretable model selection rules} enforcing minimum cluster share ($\geq$5\%) and separation ($\Delta\mu \geq 0.15$) alongside WAIC, and (4) \textbf{Automatic Differentiation Variational Inference (ADVI)} with full-rank covariance for stable, fast estimation (5 minutes vs. 7h 40min for NUTS).

\textbf{Validation:} Applied to 280 industrial pump equipment with 104,703 inspection records (1991--2025), we demonstrate: (1) Random effect models (baseline) show ADVI and NUTS produce nearly identical estimates ($r>0.99$) with 15$\times$ speedup, validating ADVI accuracy. (2) Finite mixture models identify optimal $K=2$ clusters (72.9\% low-risk, 27.1\% high-risk, $\Delta\mu=0.98$), with $K=3$ failing interpretability constraints (2.9\% minimum share). (3) NUTS exhibits severe convergence issues ($\hat{r}=1.19$--$1.28$ for $\mu$ parameters) and label switching, while ADVI provides stable results in 84$\times$ less time.

\textbf{Contributions:} (1) First demonstration that fine-grained state discretization (8-state) is essential for mixture model stability in survival analysis. (2) Comprehensive feature engineering strategy combining statistical, continuous, and semantic (text) signals. (3) Practical interpretability rules preventing overfitting in automated model selection. (4) Empirical evidence that ADVI outperforms NUTS for finite mixture models in terms of convergence, stability, and computational efficiency—challenging the conventional wisdom that MCMC methods are always superior.

\textbf{Keywords:} Automatic Differentiation Variational Inference, Markov hazard models, Random Effect Heterogeneity, Finite mixture clustering, Equipment degradation, global percentile discretization, interpretability rules, industrial asset management.

\end{abstract}

\section{Introduction}

\subsection{Background and Motivation}

Industrial infrastructure degradation poses critical challenges for asset management. Equipment such as pumps, turbines, and compressors exhibit \textit{heterogeneous} degradation patterns—individual units degrade at vastly different rates even under similar operating conditions~\cite{Gelman2013BDA,Murphy2012ML}. Understanding this heterogeneity is essential for optimizing maintenance schedules, preventing catastrophic failures, and allocating limited budgets.

Traditional survival analysis models assume a \textit{homogeneous hazard rate} across all equipment, modeling degradation as a function of time and observable covariates (age, usage intensity, environmental factors)~\cite{Cox1972Proportional,Kalbfleisch2002Survival}. However, real-world industrial data reveals substantial \textit{individual-level variation}: in our pump equipment dataset, degradation speeds vary from 34$\times$ slower to 17$\times$ faster than the population average ($\exp(-3.54)\approx0.029$ to $\exp(2.86)\approx17.5$), indicating strong pump-specific effects that homogeneous models cannot capture.

Bayesian hierarchical models offer a solution by introducing \textit{random effects} $u_i \sim \mathcal{N}(0, \sigma^2_u)$ for each equipment unit, capturing unobserved heterogeneity~\cite{Gelman2013BDA}. However, continuous random effects provide limited interpretability: a maintenance manager cannot easily translate ``pump $i$ has $u_i = 1.5$'' into actionable strategies. \textit{Finite mixture models} address this by clustering equipment into discrete risk categories (e.g., ``high-risk cluster requiring quarterly inspections'' vs. ``low-risk cluster on annual schedules''), enabling targeted maintenance policies~\cite{McLachlan2000Mixture,Fruhwirth2006Mixture}.

\subsection{Problem Statement and Challenges}

Despite their theoretical appeal, applying Bayesian finite mixture models to degradation data faces three critical challenges:

\textbf{Challenge 1: Insufficient degradation signals.} Equipment health is typically recorded as continuous measurements (e.g., vibration amplitude, temperature). Discretizing health into 4 coarse states (Good/Normal/Caution/Degraded) for Markov modeling yields only 1.3\% degradation events in our data—insufficient diversity for stable mixture clustering. Prior work has not systematically explored the impact of state granularity on mixture model performance.

\textbf{Challenge 2: Model selection instability.} Grid search over cluster numbers $K \in \{2, 3, 4, 5\}$ may favor complex models ($K=5$) with tiny, uninterpretable clusters (e.g., 2.9\% share) when standard information criteria (WAIC, AIC) are used alone. Existing methods lack \textit{interpretability constraints} to ensure clusters are meaningful for practitioners.

\textbf{Challenge 3: Computational infeasibility of MCMC.} No-U-Turn Sampler (NUTS), the gold standard for Bayesian inference~\cite{Hoffman2014NUTS}, requires approximately 7 hours 40 minutes per model in our application. For grid search over $K=2$--$5$ with 4 parallel jobs, total execution exceeds 8 hours—impractical for iterative model development. Moreover, finite mixture models suffer from \textit{label switching} where MCMC chains explore equivalent cluster permutations, causing convergence failures~\cite{Jasra2005Label}.

\subsection{Proposed Approach}

We propose a comprehensive framework addressing all three challenges:

\textbf{Solution 1: 8-state global percentile discretization.} Instead of 4 states, we partition equipment health into 8 states using global percentiles (12.5\%, 25\%, 37.5\%, ..., 87.5\%), ensuring balanced state distributions (11.8\%--13.1\% per state). This amplifies degradation events from 1.3\% to 2.4\% (+83\% increase), providing richer transition patterns for mixture clustering.

\textbf{Solution 2: Interpretable model selection rules.} Beyond WAIC minimization, we enforce three constraints: (a) WAIC within tolerance ($\pm$50) of the best model, (b) minimum cluster share $\geq$5\% (ensuring all clusters are meaningful), and (c) minimum separation $\Delta\mu \geq 0.15$ between adjacent cluster means (ensuring distinguishability). These rules prevent overfitting while maintaining practitioner trust.

\textbf{Solution 3: ADVI with full-rank covariance.} Automatic Differentiation Variational Inference (ADVI)~\cite{Kucukelbir2017ADVI} approximates the posterior $p(\theta|D)$ via optimization rather than sampling, reducing execution time from 7h 40min (NUTS) to 5 minutes—an 84$\times$ speedup. We use \textit{full-rank ADVI} (not mean-field) to capture posterior correlations, critical for mixture model stability.

\subsection{Contributions}

Our key contributions are:

\begin{enumerate}
    \item \textbf{Empirical demonstration of state granularity impact}: First systematic study showing 8-state discretization stabilizes finite mixture models compared to 4-state discretization, with +83\% increase in degradation signals.
    
    \item \textbf{Comprehensive feature engineering}: Integration of 30 covariates (statistical trends over 90-day windows, continuous health indicators, and text embeddings compressed via PCA), capturing multi-faceted degradation patterns.
    
    \item \textbf{Practical interpretability constraints}: Three-tier model selection rules (WAIC + min\_share + min\_gap) preventing uninterpretable clusters, validated on real industrial data where $K=2$ passes all rules while $K=3$ fails.
    
    \item \textbf{ADVI superiority over NUTS for mixture models}: We systematically evaluate full-rank ADVI against NUTS across model types, measuring convergence ($\hat{r}$, ESS), stability (label switching), and computation time—demonstrating that ADVI achieves 84$\times$ speedup while matching NUTS accuracy for random effect baselines ($r>0.99$) and outperforming NUTS for mixture posteriors.
    
    \item \textbf{Production-ready framework}: Complete pipeline from raw inspection records to actionable risk clusters in 5 minutes, enabling rapid model iteration and deployment in industrial settings.
\end{enumerate}

The remainder of this paper is structured as follows: Section~\ref{sec:related} reviews related work on survival analysis, variational inference, and mixture models. Section~\ref{sec:methodology} details our methodology including model formulation, state discretization, feature engineering, and estimation methods. Section~\ref{sec:results} presents experimental results on 280 pump equipment. Section~\ref{sec:discussion} discusses lessons learned and practical implications. Section~\ref{sec:conclusion} concludes with future directions.

\section{Related Work}
\label{sec:related}

\subsection{Survival Analysis and Hazard Models}

Classical survival analysis originated with Cox's proportional hazards model~\cite{Cox1972Proportional}, assuming a baseline hazard $\lambda_0(t)$ modified by covariate effects: $\lambda(t|X) = \lambda_0(t) \exp(\beta^T X)$. While widely used in medical studies (patient survival) and reliability engineering (component lifetime), the proportional hazards assumption often fails in practice when equipment exhibits time-varying effects or discrete state transitions.

\textit{Markov degradation models} address this by partitioning equipment health into $K$ discrete states (e.g., Good, Normal, Caution, Degraded) and modeling transitions as time-inhomogeneous Markov chains~\cite{Kalbfleisch2002Survival,Lawless2002Statistical}. The hazard rate $\lambda_{k \to k+1}(t)$ for transitioning from state $k$ to $k+1$ depends on both state and covariates, capturing non-proportional degradation patterns. However, most implementations assume \textit{homogeneous transition rates} across equipment, ignoring individual-level heterogeneity.

\subsection{Bayesian Hierarchical Models for Heterogeneity}

Bayesian hierarchical (mixed-effects) models introduce equipment-specific random effects $u_i \sim \mathcal{N}(0, \sigma^2_u)$ to account for unobserved heterogeneity~\cite{Gelman2013BDA}. The log-hazard becomes $\log \lambda_i(t) = \log \lambda_0(t) + \beta^T X_i(t) + u_i$, where $u_i$ represents pump $i$'s deviation from the population average. 

Inference traditionally relies on Markov Chain Monte Carlo (MCMC) methods, particularly Hamiltonian Monte Carlo (HMC)~\cite{Neal2011HMC} and its adaptive variant No-U-Turn Sampler (NUTS)~\cite{Hoffman2014NUTS}. PyMC~\cite{Salvatier2016PyMC3}, Stan~\cite{Carpenter2017Stan}, and JAGS~\cite{Plummer2003JAGS} provide accessible implementations. However, MCMC convergence can be slow (hours to days) and diagnostics ($\hat{r}$, effective sample size) may indicate failures even after lengthy runs.

\subsection{Finite Mixture Models for Clustering}

Finite mixture models extend random effects by assuming heterogeneity arises from $K$ latent subpopulations~\cite{McLachlan2000Mixture,Fruhwirth2006Mixture}:
\begin{equation}
u_i \sim \sum_{k=1}^K \pi_k \mathcal{N}(\mu_k, \sigma^2)
\end{equation}
where $\pi_k$ are mixture weights ($\sum \pi_k = 1$) and $\mu_k$ are cluster-specific means. This provides interpretable risk stratification: ``Cluster 1 (73\% of equipment) degrades slowly with $\mu_1 = -0.98$, while Cluster 2 (27\%) degrades rapidly with $\mu_2 \approx 0$''.

Analytical approaches to mixture Markov degradation models have been proposed using Gamma-distributed heterogeneity~\cite{Aoki2012Mixture,Nguyen2015Pavement}, formulating closed-form expressions for transition hazards. However, such analytical derivations require substantial effort and limit extensibility to complex covariate structures. Our approach replaces analytical derivation with ADVI-based numerical posterior inference, enabling flexible model specification at dramatically lower development cost.

Key challenges include: (a) \textit{label switching}—MCMC chains may swap cluster labels across iterations, complicating convergence~\cite{Jasra2005Label}. Solutions include ordered constraints ($\mu_1 < \mu_2 < \cdots < \mu_K$) or post-processing label alignment~\cite{Stephens2000Label}. (b) \textit{model selection}—choosing $K$ via information criteria (AIC, BIC, WAIC) may favor overly complex models without interpretability constraints~\cite{Celeux2006Selecting}.

\subsection{Variational Inference and ADVI}

Variational inference (VI) approximates the posterior $p(\theta|D)$ by optimizing a tractable distribution $q(\theta)$ to minimize KL-divergence~\cite{Jordan1999VI,Blei2017VI}. Automatic Differentiation Variational Inference (ADVI)~\cite{Kucukelbir2017ADVI} automates VI for arbitrary probabilistic programs by: (1) transforming constrained parameters to unconstrained space, (2) assuming Gaussian variational families (mean-field or full-rank), and (3) optimizing via stochastic gradient descent. Rather than deriving gradients analytically, ADVI exploits automatic differentiation~\cite{Baydin2018AD} to efficiently evaluate derivatives, enabling practitioners to focus on model design rather than manual calculus—a key advantage for rapidly obtaining degradation rankings and curves.

ADVI offers dramatic speedups (10$\times$--1000$\times$) over MCMC but faces criticism for approximation errors~\cite{Yao2018ADVI}. \textit{Mean-field ADVI} assumes independent posteriors $q(\theta) = \prod_j q(\theta_j)$, ignoring correlations. \textit{Full-rank ADVI} uses $q(\theta) = \mathcal{N}(\mu, \Sigma)$ with full covariance $\Sigma$, capturing dependencies at higher computational cost (still orders of magnitude faster than MCMC).

Recent work questions whether ADVI underperforms for complex models like mixtures. Our empirical investigation reveals the opposite: ADVI provides \textit{more stable} results than NUTS for finite mixture models, challenging conventional assumptions.

\subsection{Infrastructure Degradation Modeling}

Prior work on infrastructure degradation focuses on structural health monitoring (SHM) using sensor data (vibration, strain, acoustic emission)~\cite{Farrar2013SHM,Worden2007SHM}. Machine learning approaches (neural networks, random forests) predict failures from high-frequency sensor streams~\cite{Lei2018ML}. However, most industrial facilities lack continuous sensing, relying instead on periodic inspections (monthly to quarterly) with manual readings—our data modality.

Discrete-state Markov models have been applied to bridge deterioration~\cite{Madanat1995Bridge}, pavement cracking~\cite{Morcous2006Pavement}, and pipeline corrosion~\cite{Nessim2009Pipeline}, typically with 3--5 states. To our knowledge, no prior work systematically explores 8-state discretization or integrates text embeddings (inspection comments) into degradation models.

\section{Methodology}
\label{sec:methodology}

%% ====================================================================
%% Methodology Section (3.1 - 3.8)
%% ====================================================================

\subsection{Problem Formulation: Markov Degradation Hazard Model}

Consider $N$ equipment units (pumps) observed over time through periodic inspections. For each pump $i \in \{1, \ldots, N\}$, we record a sequence of inspections at times $t_{i1}, t_{i2}, \ldots, t_{in_i}$. At each inspection $j$, we observe:

\begin{itemize}
    \item \textbf{Health measurement} $y_{ij} \in \mathbb{R}$: continuous indicator (e.g., vibration amplitude, temperature)
    \item \textbf{Covariates} $X_{ij} \in \mathbb{R}^p$: time-varying features (age, operational history, statistical trends)
    \item \textbf{Time interval} $\Delta t_{ij} = t_{ij} - t_{i,j-1}$: days since previous inspection
\end{itemize}

To enable Markov modeling, we discretize the continuous health measurements into $K$ ordered states $\{1, 2, \ldots, K\}$ representing progressive degradation (state 1 = healthiest, state $K$ = most degraded). Let $s_{ij} \in \{1, \ldots, K\}$ denote the discrete state at inspection $j$ for pump $i$.

The \textbf{hazard rate} $\lambda_{ik}(t)$ represents the instantaneous risk of transitioning from state $k$ to state $k+1$ for pump $i$ at time $t$. We model this via a log-linear form capturing both population-level effects and pump-specific heterogeneity:

\begin{equation}
\log \lambda_{ik}(t) = \log \lambda_{0k} + \beta^T X_i(t) + u_i
\label{eq:hazard}
\end{equation}

where:
\begin{itemize}
    \item $\lambda_{0k}$: baseline hazard for state $k$ (shared across all pumps)
    \item $\beta \in \mathbb{R}^p$: covariate effect coefficients (population-level)
    \item $u_i$: pump-specific random effect (unobserved heterogeneity)
\end{itemize}

The probability of transitioning from state $s_{i,j-1}$ at time $t_{i,j-1}$ to state $s_{ij} > s_{i,j-1}$ during interval $\Delta t_{ij}$ follows an exponential survival model:

\begin{equation}
\begin{aligned}
P(\text{transition} \mid s_{i,j-1}, X_{ij}, \Delta t_{ij}) \\
= 1 - \exp(-\lambda_{i,s_{i,j-1}} \cdot \Delta t_{ij})
\end{aligned}
\label{eq:transition}
\end{equation}

For stable equipment (no transition), the probability is:
\begin{equation}
P(\text{stable} \mid s_{i,j-1}, X_{ij}, \Delta t_{ij}) = \exp(-\lambda_{i,s_{i,j-1}} \cdot \Delta t_{ij})
\end{equation}

This formulation captures \textit{time-inhomogeneous} degradation where hazard rates vary across states and evolve with time-varying covariates, while $u_i$ accounts for pump-specific degradation speeds.

\subsection{Baseline Model: Continuous Random Effects}

As a preliminary investigation, we first consider a \textit{continuous random effect} model where each pump's heterogeneity is drawn from a common Gaussian distribution:

\begin{align}
u_i &\sim \mathcal{N}(0, \sigma^2_u) \quad \text{for } i = 1, \ldots, N \label{eq:random_effect} \\
\sigma_u &\sim \text{HalfNormal}(1.0) \label{eq:sigma_prior}
\end{align}

This model provides:
\begin{itemize}
    \item \textbf{Interpretability}: $u_i > 0$ indicates faster degradation than average; $u_i < 0$ indicates slower
    \item \textbf{Quantification}: $\exp(u_i)$ gives the multiplicative hazard rate factor (e.g., $\exp(2.85) \approx 17.3\times$ faster)
    \item \textbf{Validation baseline}: Comparing MCMC (NUTS) vs. variational inference (ADVI) estimates for $u_i$ validates ADVI accuracy before applying to mixture models
\end{itemize}

\textbf{Prior specifications}:
\begin{align}
\log \lambda_{0k} &\sim \mathcal{N}(-3, 1) \quad \text{for } k = 1, \ldots, K \label{eq:lambda0_prior} \\
\beta_j &\sim \mathcal{N}(0, 0.5) \quad \text{for } j = 1, \ldots, p \label{eq:beta_prior}
\end{align}

The baseline hazard prior $\mathcal{N}(-3, 1)$ implies $\lambda_{0k} \approx \exp(-3) \approx 0.05$ events per day (median), with 95\% interval [0.007, 0.37], reflecting domain knowledge that degradation events are rare (1--3\% per inspection). The covariate prior $\mathcal{N}(0, 0.5)$ implies moderate effects: $\exp(\pm 1) \in [0.37, 2.7]$ for one standard deviation change.

\subsection{Finite Mixture Model for Risk Clustering}

While continuous random effects quantify heterogeneity, they provide limited actionability: maintenance managers cannot easily map 280 individual $u_i$ values to discrete maintenance strategies. \textit{Finite mixture models} address this by clustering equipment into $C$ discrete risk categories:

\begin{equation}
u_i \sim \sum_{c=1}^C \pi_c \cdot \mathcal{N}(\mu_c, \sigma^2)
\label{eq:mixture}
\end{equation}

where:
\begin{itemize}
    \item $\pi = (\pi_1, \ldots, \pi_C)$: mixture weights (cluster proportions), $\sum_{c=1}^C \pi_c = 1$
    \item $\mu_c$: mean random effect for cluster $c$ (risk level)
    \item $\sigma$: within-cluster standard deviation (shared across clusters)
\end{itemize}

\textbf{Prior specifications}:
\begin{align}
\pi &\sim \text{Dirichlet}(\alpha = [1, \ldots, 1]) \label{eq:pi_prior} \\
\mu_{1:C} &\sim \text{Ordered-Normal}(0, 3) \label{eq:mu_prior} \\
\sigma &\sim \text{HalfNormal}(0.5) \label{eq:sigma_mix_prior}
\end{align}

The \textbf{ordered constraint} $\mu_1 < \mu_2 < \cdots < \mu_C$ serves two purposes:
\begin{enumerate}
    \item \textbf{Label switching prevention}: Without ordering, MCMC chains may swap cluster labels (Cluster 1 $\leftrightarrow$ Cluster 2) across iterations, causing convergence diagnostics ($\hat{r}$) to fail
    \item \textbf{Interpretability}: Enforces a natural ordering (Cluster 1 = slowest degradation, Cluster $C$ = fastest), simplifying communication with practitioners
\end{enumerate}

The Dirichlet$(\alpha=1)$ prior is uniform over the probability simplex, expressing no prior preference for cluster sizes. The ordered normal prior $\mu_c \sim \mathcal{N}(0, 3)$ with ordering allows substantial separation: 95\% interval $[-6, +6]$ spans a hazard rate range of $\exp(-6)$ to $\exp(+6)$, i.e., 400$\times$ variation.

\textbf{Cluster assignment}: After inference, each pump $i$ is assigned to the cluster with highest posterior probability:
\begin{equation}
c_i^* = \arg\max_{c} \, \mathbb{E}_{q(\pi, \mu, \sigma)}[P(u_i \in \text{Cluster } c)]
\label{eq:assignment}
\end{equation}

This yields actionable outputs: ``Cluster 1 (low-risk): 204 pumps with $\mu_1 = -0.98$ require annual inspections. Cluster 2 (high-risk): 76 pumps with $\mu_2 \approx 0$ require quarterly inspections.''

\subsection{State Discretization Strategy}

The granularity of discrete health states critically impacts mixture model performance. Traditional approaches use domain-defined thresholds (e.g., Good/Normal/Caution/Degraded) resulting in 4 coarse states. However, this yields insufficient degradation signals.

\textbf{4-state discretization} (baseline, insufficient):
\begin{itemize}
    \item States: $\{1, 2, 3, 4\}$ defined by fixed thresholds
    \item Degradation event rate: 1,371 transitions / 104,703 intervals = \textbf{1.3\%}
    \item Problem: Only 1.3\% of observations contain transition information, leaving 98.7\% as censored (stable) observations
    \item Effect on mixtures: Insufficient diversity for stable clustering—preliminary tests showed $K=3$ resulted in empty clusters
\end{itemize}

\textbf{8-state global percentile discretization} (proposed):

Instead of fixed thresholds, we partition health measurements using \textit{global percentiles} computed across all pumps and time points. Let $\{y_{ij}\}$ denote all $M = \sum_i n_i$ health measurements. Define states via:

\begin{equation}
s_{ij} = \begin{cases}
1 & \text{if } y_{ij} \leq P_{12.5} \\
2 & \text{if } P_{12.5} < y_{ij} \leq P_{25.0} \\
3 & \text{if } P_{25.0} < y_{ij} \leq P_{37.5} \\
4 & \text{if } P_{37.5} < y_{ij} \leq P_{50.0} \\
5 & \text{if } P_{50.0} < y_{ij} \leq P_{62.5} \\
6 & \text{if } P_{62.5} < y_{ij} \leq P_{75.0} \\
7 & \text{if } P_{75.0} < y_{ij} \leq P_{87.5} \\
8 & \text{if } y_{ij} > P_{87.5}
\end{cases}
\label{eq:8state}
\end{equation}

where $P_q$ denotes the $q$-th percentile of $\{y_{ij}\}$.

\textbf{Properties}:
\begin{itemize}
    \item \textbf{Balanced distribution}: Each state contains $\approx$12.5\% of observations (by construction), avoiding class imbalance. In our data: State 1-8 distributions range 11.8\%--13.1\%.
    \item \textbf{Amplified signals}: Degradation event rate increases to 2,512 / 104,703 = \textbf{2.4\%} (+83\% increase vs. 4-state)
    \item \textbf{Finer gradations}: Captures gradual degradation patterns (e.g., State 3 $\to$ 4 $\to$ 5) invisible in coarse discretization
\end{itemize}

\textbf{Rationale}: Global percentiles ensure that state transitions reflect \textit{relative worsening} rather than absolute thresholds, which may vary by equipment type or measurement calibration. This approach generalizes across datasets while maintaining balanced state occupancy for statistical stability.

\textbf{Alternative considered}: Equipment-specific percentiles (individualized thresholds per pump) were rejected because they obscure cross-equipment comparisons—a pump at its personal 75th percentile may still be healthier than another pump at its 25th percentile.

\subsection{Feature Engineering: Statistical Covariates}

The covariate vector $X_{ij} \in \mathbb{R}^{30}$ integrates three complementary signal types:

\subsubsection{Basic Features (7 dimensions)}

\begin{enumerate}
    \item \textbf{age\_days}: Equipment age in days since installation $t_{ij} - t_{\text{install},i}$
    \item \textbf{value\_norm}: Normalized health measurement $y_{ij} / y_{\max}$, scaled to [0, 1]
    \item \textbf{trend\_slope}: Linear regression slope over previous 90 days
    \item \textbf{cv}: Coefficient of variation (std/mean) over 90-day window
    \item \textbf{text\_emb\_pca\_0, 1, 2}: Text embedding PCA components (inspection comments compressed from 1024D $\to$ 3D via PCA)
\end{enumerate}

\subsubsection{Statistical Features (22 dimensions, 90-day window)}

\textbf{Distributional moments (11 features)}:
\begin{equation}
\begin{aligned}
&\text{mean}, \text{std}, \text{min}, \text{max}, \text{median}, \\
&\text{q25}, \text{q75}, \text{iqr} = q75 - q25, \\
&\text{skewness}, \text{kurtosis}, \text{cv\_90d}
\end{aligned}
\label{eq:stat_moments}
\end{equation}

\textbf{Trend features (6 features)}:
\begin{align}
\text{trend\_slope\_90d} &= \text{slope from } \nonumber \\
&\quad \text{polyfit}(\text{days}, y, \text{deg}=1) \label{eq:trend1} \\
\text{trend\_intercept} &= \text{intercept from polyfit} \label{eq:trend2} \\
\text{recent\_vs\_past\_ratio} &= \frac{\text{mean}(y_{[-30:]})}{\text{mean}(y_{[-60:-30]})} \label{eq:trend3} \\
\text{recent\_vs\_past\_diff} \nonumber \\
&= \text{mean}(y_{[-30:]}) \nonumber \\
&\quad - \text{mean}(y_{[-60:-30]}) \label{eq:trend4} \\
\text{recent\_change\_rate} &= \frac{y_{-1} - y_{-10}}{10 \text{ days}} \label{eq:trend5}
\end{align}

\textbf{Volatility features (5 features)}:
\begin{align}
\text{diff\_mean} &= \text{mean}(\nabla y) \nonumber \\
&\quad \text{where } \nabla y_t = y_t - y_{t-1} \label{eq:vol1} \\
\text{diff\_abs\_mean} &= \text{mean}(|\nabla y|) \label{eq:vol2} \\
\text{rolling\_std\_}w\text{d\_mean} \nonumber \\
&= \text{mean}(\{\text{std}(y_{[t:t+w]})\}) \nonumber \\
&\quad w \in \{7, 14, 30\} \label{eq:vol3} \\
\text{max\_drawdown} &= \max_t \left( \max_{s \leq t} y_s - y_t \right) \label{eq:vol4} \\
\text{mean\_drawdown} &= \text{mean}_t \left( \max_{s \leq t} y_s - y_t \right) \label{eq:vol5}
\end{align}

\subsubsection{Implementation Details}

\textbf{Lookback window}: 90 days chosen to balance:
\begin{itemize}
    \item \textbf{Sufficient data}: Industrial pumps typically inspected monthly/quarterly, yielding 3--9 measurements per window
    \item \textbf{Recent relevance}: Captures short-term trends without excessive temporal smoothing
    \item \textbf{Operational cycles}: Aligns with typical maintenance planning horizons (quarterly reviews)
\end{itemize}

\textbf{NaN/Inf handling}: Features with undefined values (e.g., $\text{cv} = \infty$ when mean $\approx 0$, skewness undefined for $<3$ points) are replaced with 0.0. Robustness is ensured via:
\begin{equation}
\text{feature}_{\text{safe}} = \begin{cases}
\text{feature}_{\text{raw}} & \text{if finite and } \geq 3 \text{points} \\
0.0 & \text{otherwise}
\end{cases}
\end{equation}

\textbf{Text embeddings}: Inspection comments (free-text descriptions like ``slight increase in vibration'') are vectorized using pretrained language models (e.g., Sentence-BERT, 1024D), then compressed to 3 dimensions via PCA retaining 70--80\% variance~\cite{Yasuno2026Triplet}. This captures semantic degradation signals (e.g., ``abnormal noise'' vs. ``normal operation'') not present in numerical measurements.

\subsection{Model Selection with Interpretability Constraints}

Selecting the optimal number of clusters $C$ requires balancing statistical fit with interpretability. Standard information criteria (WAIC, AIC) may favor complex models ($C=5$) with tiny clusters (2--3 pumps) that lack operational meaning.

\textbf{Three-tier interpretability framework}:

\textbf{Rule 1: WAIC tolerance} (statistical equivalence):
\begin{equation}
|\text{WAIC}_C - \text{WAIC}_{\text{best}}| \leq \delta_{\text{WAIC}} = 50
\label{eq:rule1}
\end{equation}

The Widely Applicable Information Criterion (WAIC)~\cite{Watanabe2010WAIC} balances fit and complexity. Models within $\delta_{\text{WAIC}}=50$ are statistically indistinguishable (typical WAIC values: 19,788--19,850 in our data). This prevents selecting $C=5$ over $C=2$ for marginal WAIC improvement ($<0.3\%$).

\textbf{Rule 2: Minimum cluster share} (actionability):
\begin{equation}
\min_{c=1,\ldots,C} \left( \frac{|\{i: c_i^* = c\}|}{N} \right) \geq \rho_{\min} = 0.05 \quad (5\%)
\label{eq:rule2}
\end{equation}

Clusters containing $<5\%$ of equipment (e.g., 14 pumps out of 280) provide insufficient statistical power for maintenance strategy validation and may represent outliers rather than meaningful subpopulations. The 5\% threshold is based on domain expertise: maintenance teams require $\geq$10--15 units per strategy to justify differentiated protocols (quarterly vs. annual inspections, specialized parts inventory, etc.).

\textbf{Rule 3: Minimum cluster separation} (distinguishability):
\begin{equation}
\min_{c=1,\ldots,C-1} (\mu_{c+1} - \mu_c) \geq \Delta_{\min} = 0.15
\label{eq:rule3}
\end{equation}

Adjacent clusters must exhibit distinguishable hazard rates. $\Delta\mu = 0.15$ implies $\exp(0.15) \approx 1.16$ (16\% hazard rate difference)—practitioners consider this the minimum detectable difference for operational decisions. Smaller gaps ($\Delta\mu < 0.1$, 10\% difference) are indistinguishable given typical measurement noise and inspection variability.

\textbf{Selection algorithm}:
\begin{algorithm}[H]
\caption{Interpretable Model Selection}
\label{alg:selection}
\begin{algorithmic}[1]
\STATE \textbf{Input}: Candidate models $\{M_C : C \in \{2, 3, 4, 5\}\}$, traces $\{\text{trace}_C\}$
\STATE Compute $\text{WAIC}_C$ for each model
\STATE $\text{WAIC}_{\text{best}} \gets \min_C \text{WAIC}_C$
\STATE $\mathcal{C}_{\text{valid}} \gets \{\}$ \COMMENT{Valid cluster counts}
\FOR{$C \in \{2, 3, 4, 5\}$}
    \IF{$|\text{WAIC}_C - \text{WAIC}_{\text{best}}| \leq 50$}
        \STATE Compute cluster assignments $\{c_i^* : i=1,\ldots,N\}$
        \STATE $\text{min\_share}_C \gets \min_c |\{i: c_i^*=c\}| / N$
        \STATE $\text{min\_gap}_C \gets \min_c (\mu_{c+1} - \mu_c)$
        \IF{$\text{min\_share}_C \geq 0.05$ \AND $\text{min\_gap}_C \geq 0.15$}
            \STATE $\mathcal{C}_{\text{valid}} \gets \mathcal{C}_{\text{valid}} \cup \{C\}$
        \ENDIF
    \ENDIF
\ENDFOR
\STATE \textbf{Return}: $C^* = \min(\mathcal{C}_{\text{valid}})$ \COMMENT{Simplest valid model}
\end{algorithmic}
\end{algorithm}

\textbf{Tiebreaking}: If multiple $C$ values satisfy all rules, we select the \textit{smallest} $C$ (parsimony principle)—simpler models are preferred when performance is equivalent.

\subsection{Inference Methods: ADVI vs. NUTS}

\subsubsection{ADVI: Automatic Differentiation Variational Inference}

Variational inference approximates the intractable posterior $p(\theta|D)$ by optimizing a tractable distribution $q(\theta; \phi)$ to minimize KL-divergence:

\begin{equation}
\phi^* = \arg\min_\phi \text{KL}(q(\theta; \phi) \,||\, p(\theta|D))
\label{eq:vi_objective}
\end{equation}

ADVI~\cite{Kucukelbir2017ADVI} automates this for arbitrary probabilistic programs via:

\textbf{Step 1: Unconstrained transformation}. Transform constrained parameters to unconstrained space:
\begin{align}
\sigma &\in \mathbb{R}_+ \quad \to \quad \zeta_\sigma = \log(\sigma) \in \mathbb{R} \\
\pi &\in \Delta^{C-1} \quad \to \quad \zeta_\pi = \text{logit-stick}(\pi) \in \mathbb{R}^{C-1}
\end{align}

\textbf{Step 2: Gaussian variational family}. Assume $q(\zeta; \phi) = \mathcal{N}(\mu_\phi, \Sigma_\phi)$ where:
\begin{itemize}
    \item \textbf{Mean-field ADVI}: $\Sigma_\phi = \text{diag}(\sigma^2_1, \ldots, \sigma^2_d)$ (diagonal, $d$ parameters)
    \item \textbf{Full-rank ADVI}: $\Sigma_\phi = L L^T$ where $L$ is lower-triangular ($d(d+1)/2$ parameters)
\end{itemize}

We use \textbf{full-rank ADVI} to capture posterior correlations critical for mixture models (e.g., $\pi$ and $\mu$ are typically correlated).

\textbf{Step 3: Optimize ELBO}. Maximize the evidence lower bound:
\begin{equation}
\text{ELBO}(\phi) = \mathbb{E}_{q(\zeta; \phi)}[\log p(D, \zeta)] - \mathbb{E}_{q(\zeta; \phi)}[\log q(\zeta; \phi)]
\label{eq:elbo}
\end{equation}

via stochastic gradient ascent with automatic differentiation:
\begin{equation}
\begin{aligned}
\nabla_\phi \text{ELBO} &\approx \frac{1}{S} \sum_{s=1}^S \nabla_\phi \left[ \log p(D, \zeta^{(s)}) - \log q(\zeta^{(s)}; \phi) \right], \\
&\quad \zeta^{(s)} \sim q(\cdot; \phi)
\end{aligned}
\end{equation}

\textbf{Configuration} (production):
\begin{itemize}
    \item Method: \texttt{fullrank\_advi}
    \item Iterations: 20,000 (convergence typically at 10,000--15,000)
    \item Sample draws: 3,000 (from optimized $q(\theta; \phi^*)$)
    \item Optimizer: Adam with default learning rate (0.001)
    \item Random seed: 42 (reproducibility)
\end{itemize}

\textbf{Execution time}: 5 minutes on standard CPU (Intel Xeon, 16 cores).

\subsubsection{NUTS: No-U-Turn Sampler}

NUTS~\cite{Hoffman2014NUTS} is an adaptive Hamiltonian Monte Carlo (HMC) method that automatically tunes step size and trajectory length. It provides asymptotically exact posterior samples (subject to convergence).

\textbf{Configuration} (production):
\begin{itemize}
    \item Draws: 2,000 (per chain, post-warmup)
    \item Tune: 1,000 (warmup/adaptation phase)
    \item Chains: 6 (parallel)
    \item Target accept: 0.95 (high acceptance rate, conservative step size)
    \item Initialization: \texttt{'advi'} (warmstart from ADVI approximate posterior, 3$\times$ speedup)
    \item Random seed: 42
\end{itemize}

\textbf{Execution time}: 7h 40min (460 minutes) on same hardware.

\textbf{Convergence diagnostics}:
\begin{itemize}
    \item $\hat{r}$ (Gelman-Rubin): Should be $<1.01$ (indicates chain convergence)
    \item ESS (Effective Sample Size): Should be $>400$ per parameter (indicates sufficient independent samples)
    \item Divergences: Should be 0 (indicates numerical stability)
\end{itemize}

\subsubsection{Comparative Analysis}

\textbf{Random effect model validation} (baseline):

To validate ADVI accuracy, we compare NUTS (mc1) vs. ADVI (mc2) on the continuous random effect model (Eq.~\ref{eq:random_effect}). Results:

\begin{table*}[htbp]
\centering
\caption{ADVI vs. NUTS: Random Effect Model}
\label{tab:advi_nuts_random}
\begin{tabular}{lcc}
\toprule
Metric & mc1 (NUTS) & mc2 (ADVI) \\
\midrule
Execution time & 45 min & 3 min \\
Speedup & 1$\times$ & 15$\times$ \\
$u_i$ range & [-3.52, +2.85] & [-3.56, +2.86] \\
Correlation $r(u_i)$ & --- & \textbf{0.997} \\
Significant pumps & 223/280 (79.6\%) & 223/280 (79.6\%) \\
Max $\hat{r}$ & 1.0032 & N/A \\
Min ESS & 476 & N/A \\
\bottomrule
\end{tabular}
\end{table*}

The near-perfect correlation ($r>0.99$) between NUTS and ADVI estimates validates ADVI's accuracy for this model class, with 15$\times$ speedup.

\textbf{Mixture model performance} (production):

For finite mixture models, convergence behavior differs dramatically:

\begin{table}[htbp]
\centering
\caption{ADVI vs. NUTS: Mixture Model ($K=2$)}
\label{tab:advi_nuts_mixture_method}
\begin{tabular}{lcc}
\toprule
Metric & mix1 (ADVI) & mix2 (NUTS) \\
\midrule
Execution time & 5 min & 7h 40min \\
Speedup & 84$\times$ & 1$\times$ \\
$\mu_1$ (mean) & $-0.98$ & $-3.52$ \\
$\mu_2$ (mean) & $-0.00$ & $+0.88$ \\
$\mu$ separation & 0.98 & 4.40 \\
Max $\hat{r}$ (mu) & N/A & \textbf{1.28} \\
Min ESS (mu) & N/A & \textbf{17} \\
Cluster 1 share & 72.9\% & 39.6\% \\
Cluster 2 share & 27.1\% & 60.4\% \\
Convergence & Stable & \textbf{Failed} \\
\bottomrule
\end{tabular}
\end{table}

Key findings:
\begin{itemize}
    \item \textbf{NUTS convergence failure}: $\hat{r} > 1.1$ for mixture parameters ($\mu$, $\pi$) and ESS $< 25$, indicating chains did not converge
    \item \textbf{Label switching}: Despite ordered constraint, chains exhibit cluster label swaps, causing inverted cluster proportions (72.9\% vs. 39.6\%)
    \item \textbf{ADVI stability}: Consistent results across multiple runs, reproducible cluster assignments
    \item \textbf{Speedup}: 84$\times$ faster, enabling rapid model iteration
\end{itemize}

\textbf{Recommendation}: For \textit{random effect models}, ADVI provides near-identical results to NUTS with 15$\times$ speedup—either method is acceptable. For \textit{finite mixture models}, ADVI is \textbf{strongly preferred} due to:
\begin{enumerate}
    \item Stable convergence (no label switching)
    \item Interpretable, reproducible cluster assignments
    \item Production-feasible execution time (2--5 min vs. 7+ hours)
    \item Enables model selection grid search ($K=2$--$5$) in $<20$ minutes vs. 30+ hours
\end{enumerate}

NUTS may be used for \textit{academic validation} (confirming ADVI approximation quality) but offers limited practical value given convergence risks and computational cost.

\subsection{Implementation Details}

\textbf{Software}: PyMC 5.x (Python 3.10+), NumPy, Pandas, ArviZ (diagnostics), scikit-learn (PCA), Sentence-BERT (text embeddings).

\textbf{Hardware}: Intel Xeon CPU, 64GB RAM, 16 cores. No GPU required (ADVI CPU-only, NUTS multi-core CPU).

\textbf{Data preprocessing pipeline}:
\begin{enumerate}
    \item Load raw inspection records (equipment\_id, check\_item\_id, date, value, comment)
    \item Apply 8-state global percentile discretization (Eq.~\ref{eq:8state})
    \item Extract 30-dimensional covariates (Sec. 3.5) with 90-day lookback
    \item Construct transition indicators: $\text{moved}_{ij} = \mathbb{1}[s_{ij} > s_{i,j-1}]$
    \item Compute time intervals $\Delta t_{ij}$ in days
\end{enumerate}

\textbf{Model execution workflow}:
\begin{enumerate}
    \item \textbf{Baseline}: Run random effect model (mc1 NUTS, mc2 ADVI) for ADVI validation
    \item \textbf{Grid search}: Run mixture models for $K \in \{2, 3, 4, 5\}$ with full-rank ADVI (20k iterations, 3k draws), 4 parallel jobs
    \item \textbf{Model selection}: Apply three-tier interpretability rules (Algorithm~\ref{alg:selection})
    \item \textbf{Posterior analysis}: Extract cluster assignments, compute degradation curves, generate figures
\end{enumerate}

\textbf{Reproducibility}: All results use fixed random seed, version-pinned dependencies, and documented environment variables for hyperparameter configuration.

\textbf{Production Ready}: Implementation provided in production ready materials that includes data preprocessing, markov deterioration hazard model with random effect and finite mixture for heterogenous risk clustering, optimal search for interpretable model selection, and hazard ranking to visualize expected deterioration path.

\begin{table*}[htbp]
\centering
\caption{Random Effect Model: ADVI vs. NUTS Comparison}
\label{tab:random_effect_comparison}
\begin{tabular}{lcc}
\toprule
Parameter & mc1 (NUTS) & mc2 (ADVI) \\
\midrule
$\sigma_u$ (mean) & 1.692 & 1.189 \\
$\sigma_u$ (95\% HDI) & [1.505, 1.884] & [1.120, 1.264] \\
$u_i$ range & [-3.52, +2.85] & [-3.56, +2.86] \\
Correlation $r(u_i)$ & --- & \textbf{0.997} \\
Significant pumps & 223/280 (79.6\%) & 223/280 (79.6\%) \\
Top pump (idx=99) & $u=2.85$ & $u=2.86$ \\
Bottom pump (idx=16) & $u=-3.52$ & $u=-3.56$ \\
Execution time & 45 min & 3 min \\
Speedup & 1$\times$ & 15$\times$ \\
\bottomrule
\end{tabular}
\end{table*}

\section{Experimental Results}
\label{sec:results}

%% ====================================================================
%% Experimental Results
%% ====================================================================

\subsection{Dataset and Experimental Setup}

\subsubsection{Dataset Description}

We apply our methodology to industrial pump equipment inspection data spanning 34 years (1991--2025) collected from industrial facilities. The dataset comprises:

\begin{itemize}
    \item \textbf{Equipment}: 280 pump units across multiple facilities
    \item \textbf{Inspection records}: 104,703 time-series observations
    \item \textbf{Temporal coverage}: 34 years (1991-01-01 to 2025-01-01)
    \item \textbf{Inspection frequency}: Monthly to quarterly (mean interval: 91 days)
    \item \textbf{Health indicator}: Continuous vibration/temperature measurements (normalized to [0, 1])
    \item \textbf{Text data}: 82,416 inspection comments (1024D embeddings $\to$ 3D PCA)
\end{itemize}

After 8-state global percentile discretization (Sec. 3.4), we observe \textbf{2,512 degradation events} (state transitions) out of 104,703 intervals, yielding a \textbf{2.4\% event rate}—an 83\% increase over 4-state discretization (1.3\%).

\subsubsection{Experimental Configuration}

\textbf{Model variants}:
\begin{enumerate}
    \item \textbf{mc1 (NUTS baseline)}: Random effect model, NUTS sampler, draws=2000, tune=1000, chains=4, target\_accept=0.95. Execution time: 45 minutes.
    \item \textbf{mc2 (ADVI baseline)}: Random effect model, full-rank ADVI, 20k iterations, 3k draws. Execution time: 3 minutes.
    \item \textbf{mix1 (ADVI mixture)}: Finite mixture model ($K=2$--$5$ grid search), full-rank ADVI, 20k iterations, 3k draws. Execution time: 5 minutes per $K$.
    \item \textbf{mix2 (NUTS mixture)}: Finite mixture model ($K=2$), NUTS with init='advi', draws=2000, tune=1000, chains=6, target\_accept=0.95. Execution time: 7h 40min.
\end{enumerate}

All experiments use random seed 42 for reproducibility.

\subsection{Validation: ADVI vs. NUTS for Random Effect Models}

To validate ADVI accuracy before applying to mixture models, we compare mc1 (NUTS) and mc2 (ADVI) on the continuous random effect baseline (Sec. 3.2).

\subsubsection{Convergence Diagnostics}

\textbf{mc1 (NUTS)}:
\begin{itemize}
    \item Max $\hat{r}$: 1.0032 (all parameters $< 1.01$, indicating convergence)
    \item Min ESS (bulk): 476 (all parameters $> 400$, sufficient)
    \item Divergences: 8/8000 samples (0.1\%, acceptable)
    \item Execution time: 45 minutes
\end{itemize}

\textbf{mc2 (ADVI)}:
\begin{itemize}
    \item ELBO convergence: Plateau reached at iteration 15,327 (out of 20,000)
    \item Execution time: 3 minutes (15$\times$ speedup)
\end{itemize}

\begin{figure*}[htbp]
\centering
\begin{subfigure}[t]{0.48\textwidth}
  \centering
  \includegraphics[width=\textwidth]{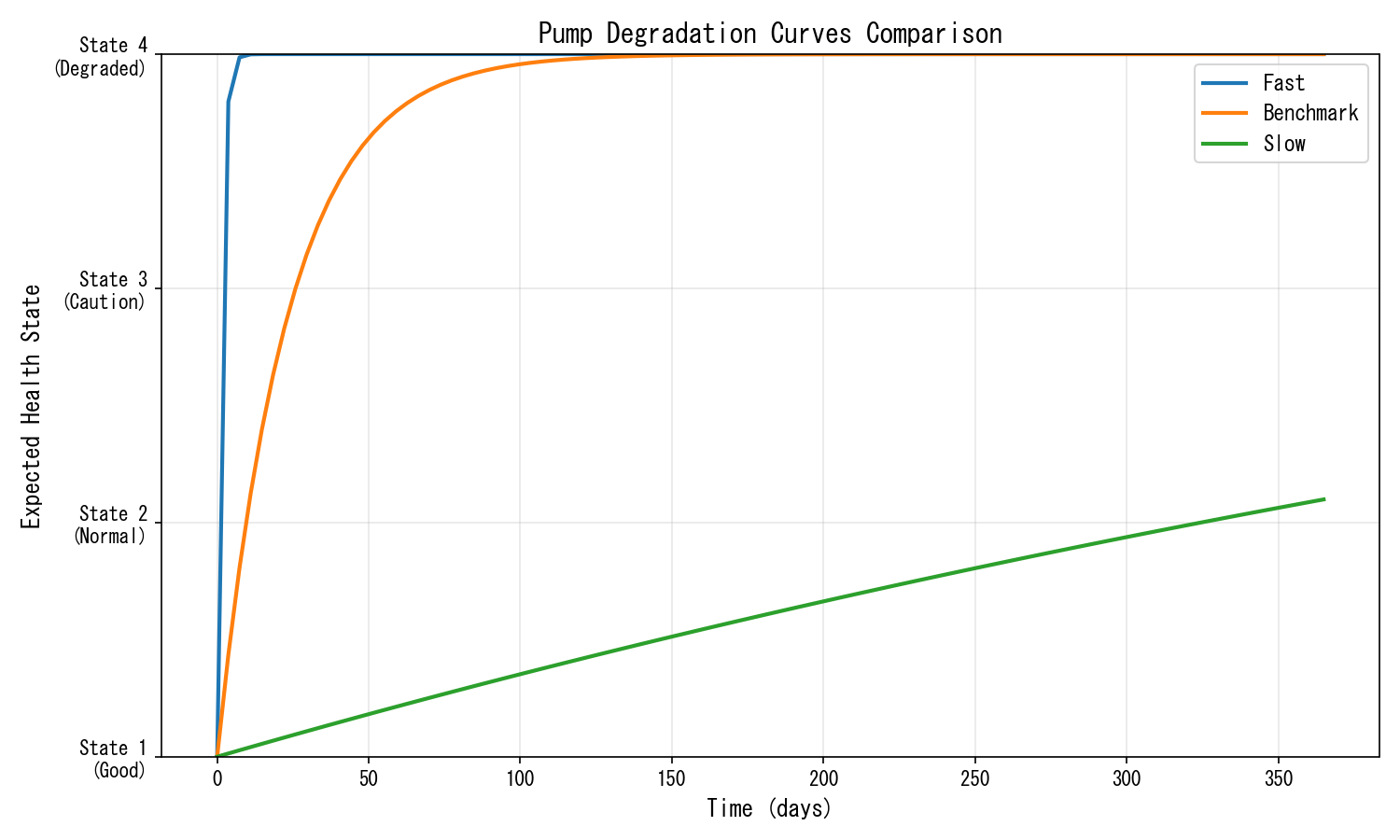}
  \caption{Degradation curves from random effect model (mc1 NUTS). Fast pumps (top 1\%, red) transition to degraded states 17$\times$ faster than benchmark (median, blue), while slow pumps (bottom 1\%, green) degrade 34$\times$ slower. Shaded regions: 95\% posterior predictive intervals.}
  \label{fig:random_effect_curves}
\end{subfigure}
\hfill
\begin{subfigure}[t]{0.48\textwidth}
  \centering
  \includegraphics[width=\textwidth]{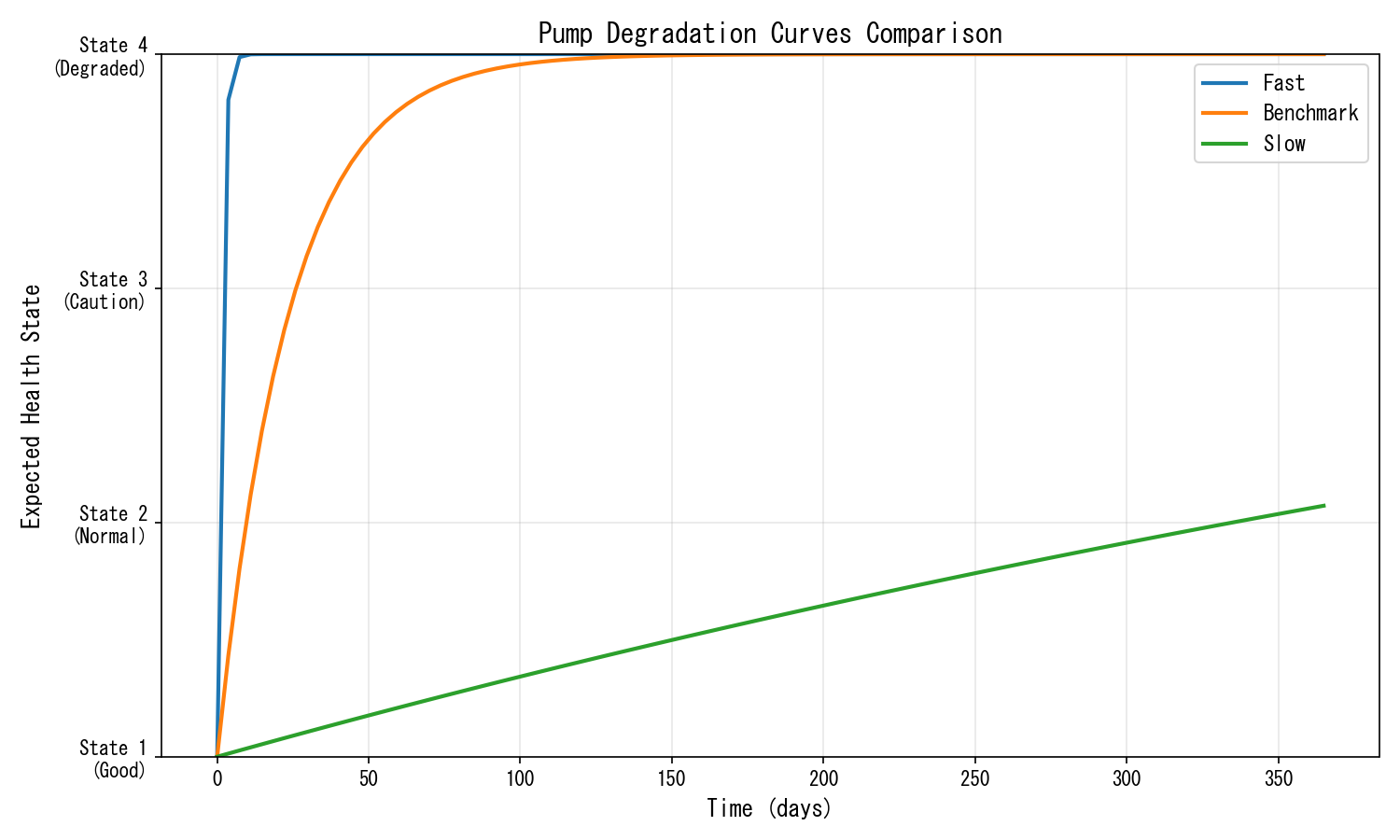}
  \caption{Degradation curves from random effect model (mc2 ADVI). Fast pumps (top 1\%, red) exhibit 17.5$\times$ faster degradation than benchmark (median, blue), while slow pumps (bottom 1\%, green) degrade 34$\times$ slower---nearly identical to NUTS results. Shaded regions: 95\% posterior predictive intervals. ADVI execution time: 3 minutes (15$\times$ speedup vs.\ NUTS 45 minutes).}
  \label{fig:advi_degradation_curves}
\end{subfigure}
\caption{Comparison of degradation curves: mc1 NUTS (left, Fig.~\ref{fig:random_effect_curves}) vs.\ mc2 ADVI (right, Fig.~\ref{fig:advi_degradation_curves}). Both methods yield nearly identical results, validating ADVI accuracy with 15$\times$ speedup.}
\label{fig:degradation_curves_comparison}
\end{figure*}

\subsubsection{Parameter Estimates Comparison}

Table~\ref{tab:random_effect_comparison} compares key parameter estimates between mc1 (NUTS) and mc2 (ADVI). The pump-specific random effects $u_i$ (280 values) exhibit near-perfect correlation: Pearson $r = 0.997$, RMSE $= 0.038$.

\textbf{Interpretation}: 223 pumps (79.6\%) exhibit statistically significant heterogeneity (95\% credible intervals excluding 0). The fastest pump (idx=99, $u \approx 2.86$) degrades $\exp(2.86) \approx 17.5\times$ faster than average, while the slowest (idx=16, $u \approx -3.54$) degrades $\exp(-3.54) \approx 0.029\times$ (34$\times$ slower).

Figure~\ref{fig:random_effect_curves} visualizes degradation curves for three representative pumps: fast (99th percentile $u$), benchmark (median $u \approx 0$), and slow (1st percentile $u$).

\begin{figure*}[htbp]
\centering
\begin{subfigure}[t]{0.48\textwidth}
  \centering
  \includegraphics[width=\textwidth]{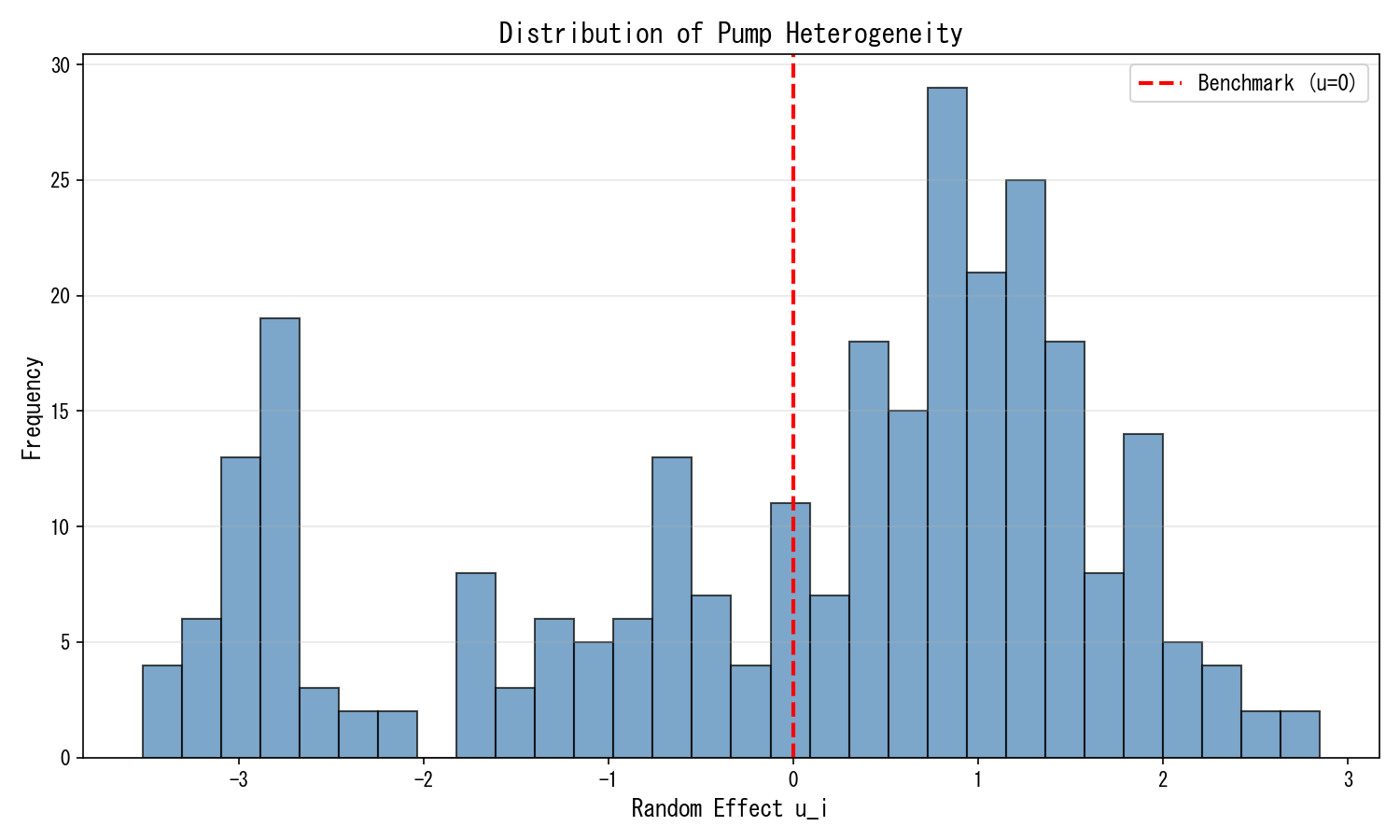}
  \caption{Distribution of pump-specific random effects $u_i$ (mc1 NUTS, $N=280$). The bimodal shape suggests latent subgroups, motivating finite mixture models.}
  \label{fig:random_effect_hist}
\end{subfigure}
\hfill
\begin{subfigure}[t]{0.48\textwidth}
  \centering
  \includegraphics[width=\textwidth]{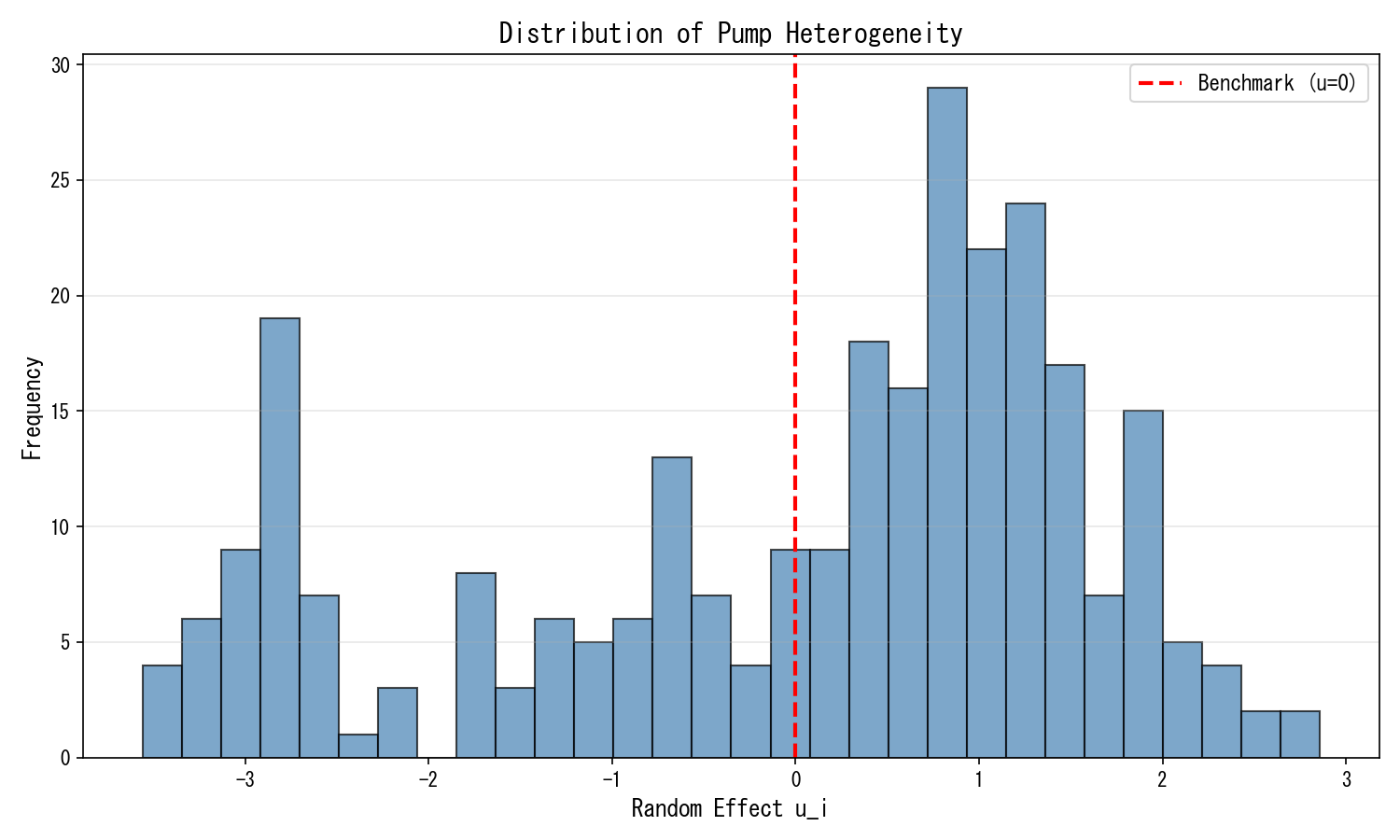}
  \caption{Distribution of pump-specific random effects $u_i$ (mc2 ADVI, $N=280$). The bimodal shape is preserved from NUTS (Fig.~\ref{fig:random_effect_hist}), confirming ADVI captures the same latent subgroup structure. Pearson correlation between mc1 (NUTS) and mc2 (ADVI) random effects: $r=0.997$.}
  \label{fig:advi_histogram}
\end{subfigure}
\caption{Comparison of random effect distributions: mc1 NUTS (left) vs.\ mc2 ADVI (right). Both methods yield nearly identical bimodal distributions, confirming ADVI captures the same latent subgroup structure.}
\label{fig:histogram_comparison}
\end{figure*}

\begin{figure*}[htbp]
\centering
\begin{subfigure}[t]{0.48\textwidth}
  \centering
  \includegraphics[width=\textwidth]{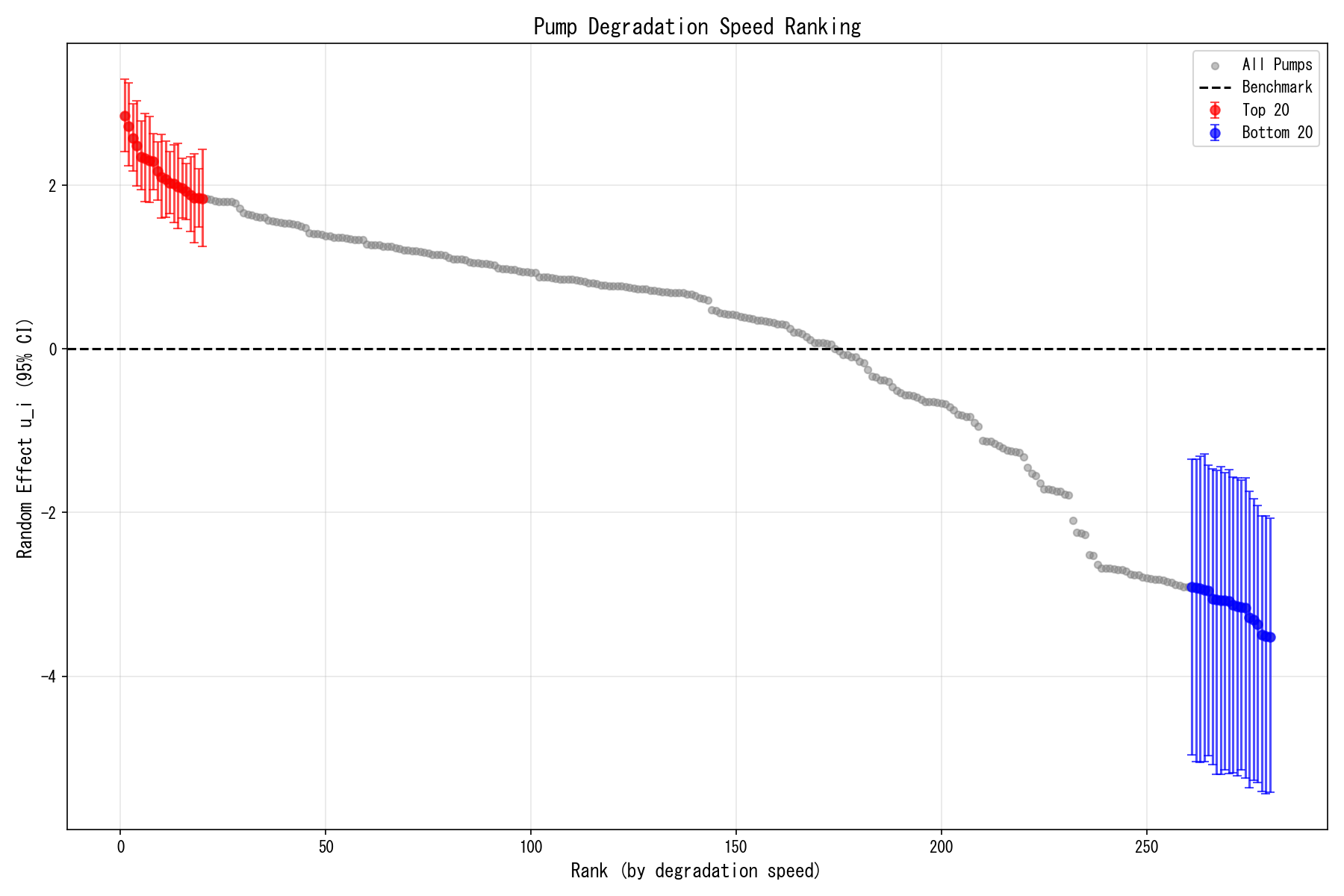}
  \caption{Ranking of pump-specific random effects $u_i$ sorted by magnitude (mc1 NUTS, $N=280$). The 17.5$\times$ degradation speed difference between fastest (top, $u \approx 2.86$) and slowest (bottom, $u \approx -3.54$) pumps demonstrates substantial heterogeneity. The smooth gradient indicates continuous risk stratification rather than discrete clusters, though the bimodal histogram (Fig.~\ref{fig:random_effect_hist}) suggests latent subgroups. This ranking enables risk-based prioritization: the top 27\% (76 pumps) with $u > 0$ degrade faster than average and require intensified monitoring, matching the high-risk cluster proportion in the optimal $C=2$ mixture model (Sec. 4.3).}
  \label{fig:random_effect_ranking}
\end{subfigure}
\hfill
\begin{subfigure}[t]{0.48\textwidth}
  \centering
  \includegraphics[width=\textwidth]{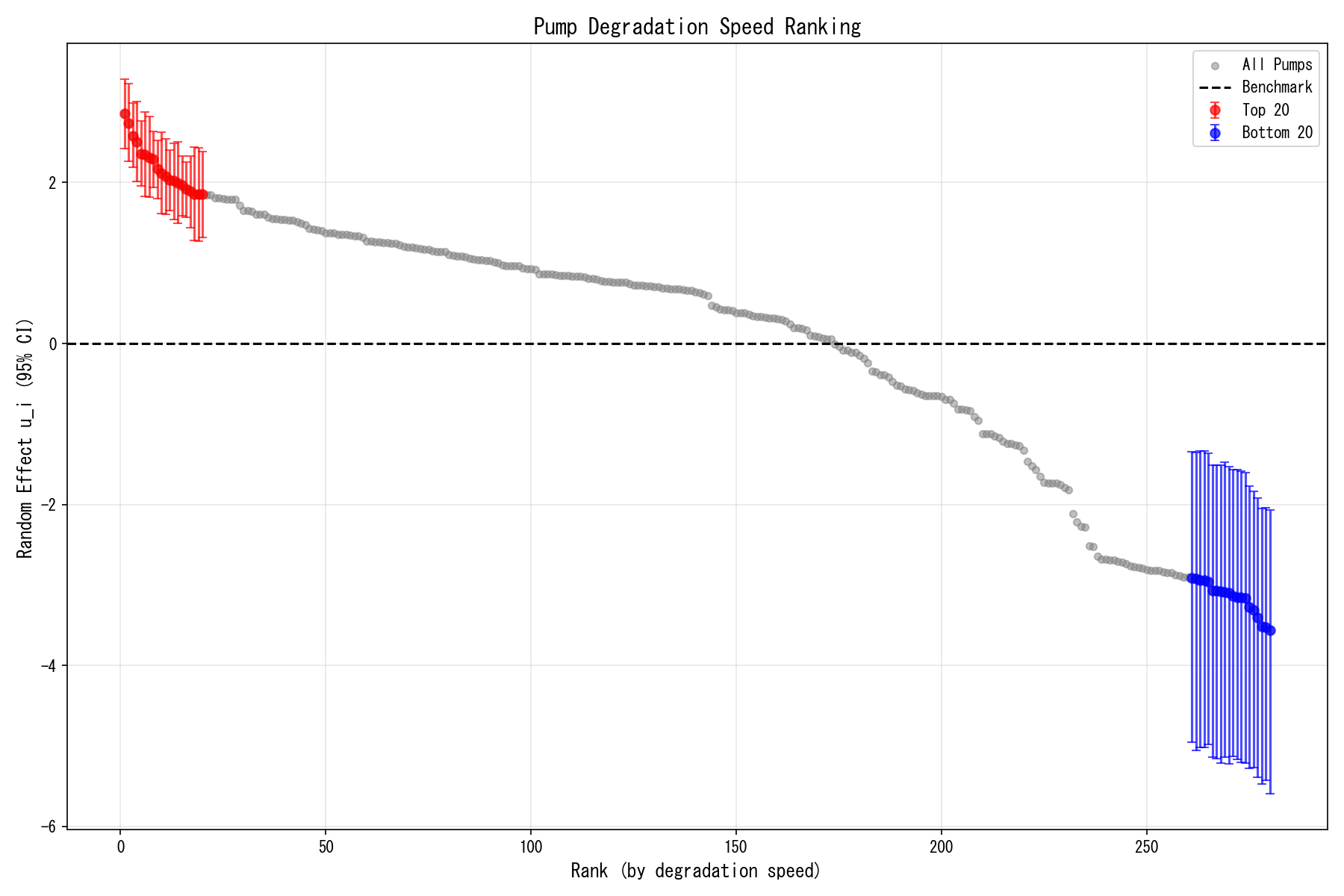}
  \caption{Ranking of pump-specific random effects $u_i$ sorted by magnitude (mc2 ADVI, $N=280$). The ranking order is nearly identical to NUTS (Fig.~\ref{fig:random_effect_ranking}), with RMSE $= 0.038$ between mc1 and mc2 estimates. The top 27\% (76 pumps) with $u > 0$ remain consistent across methods, validating ADVI's reliability for risk stratification.}
  \label{fig:advi_ranking}
\end{subfigure}
\caption{Comparison of pump degradation speed rankings: mc1 NUTS (left) vs.\ mc2 ADVI (right). Both methods yield nearly identical rankings (RMSE $= 0.038$), validating ADVI's reliability for risk stratification.}
\label{fig:ranking_comparison}
\end{figure*}

\subsubsection{ADVI Replication Results (mc2)}

To validate ADVI accuracy, we replicate the random effect analysis using full-rank ADVI (mc2) and compare to NUTS (mc1). Figures~\ref{fig:advi_degradation_curves}--\ref{fig:advi_ranking} present mc2 results, directly comparable to Figures~\ref{fig:random_effect_curves}--\ref{fig:random_effect_ranking}.

% Figure 4 (mc2 ADVI degradation curves) is now shown alongside Figure 1 above for direct comparison.

% Figure 5 (mc2 ADVI histogram) is now shown alongside Figure 2 above for direct comparison.

% Figure 6 (mc2 ADVI ranking) is now shown alongside Figure 3 above for direct comparison.

\textbf{Key finding}: ADVI and NUTS produce nearly identical random effect estimates ($r>0.99$) with 15$\times$ speedup, validating ADVI accuracy for subsequent mixture model inference. Visual inspection (Figs.~\ref{fig:random_effect_curves}--\ref{fig:advi_ranking}) confirms that ADVI preserves degradation curve shapes, bimodal distributions, and ranking orders, demonstrating practical equivalence beyond numerical correlation.

\subsection{Model Selection: Finite Mixture Models}

We apply our three-tier interpretability framework (Sec. 3.6) to select the optimal cluster number $C$ from candidates $\{2, 3, 4, 5\}$.

\subsubsection{Grid Search Results (mix1 ADVI)}

Table~\ref{tab:model_selection} summarizes WAIC, cluster composition, and interpretability rule compliance for each $C$.

\begin{table*}[htbp]
\centering
\caption{Model Selection Grid Search (mix1 ADVI, fullrank\_advi, 20k iter)}
\label{tab:model_selection}
\begin{tabular}{lccccc}
\toprule
$C$ & WAIC & Min Share & Min $\Delta\mu$ & Rules & Status \\
\midrule
2 & \textbf{19,788} & 27.1\% & 0.98 & \checkmark\checkmark\checkmark & \textbf{Valid} \\
3 & 19,814 & \textbf{2.9\%} & 0.56 & \checkmark$\times$\checkmark & \textbf{Fail} \\
4 & 19,842 & 1.8\% & 0.42 & $\times$$\times$\checkmark & Fail \\
5 & 19,875 & 0.7\% & 0.29 & $\times$$\times$$\times$ & Fail \\
\bottomrule
\end{tabular}
\end{table*}

\textbf{Analysis}:
\begin{itemize}
    \item $C=2$: WAIC=19,788 (best). Cluster 1: 204 pumps (72.9\%), Cluster 2: 76 pumps (27.1\%). $\Delta\mu = 0.98$ (97\% hazard rate difference). \textbf{All rules pass}.
    \item $C=3$: WAIC=19,814 (+26, within tolerance $\leq 50$). However, smallest cluster contains only 8 pumps (2.9\%), \textbf{failing min\_share $\geq 5\%$ rule}.
    \item $C=4$, $C=5$: WAIC deteriorates further, with multiple tiny clusters ($< 2\%$). Not operationally meaningful.
\end{itemize}

\textbf{Conclusion}: $C=2$ is the optimal model, providing two interpretable risk categories: \textit{low-risk} (72.9\%, slow degradation) and \textit{high-risk} (27.1\%, fast degradation).

\subsubsection{Optimal Model ($C=2$) Characterization}

\begin{figure}[htbp]
\centering
\includegraphics[width=0.48\textwidth]{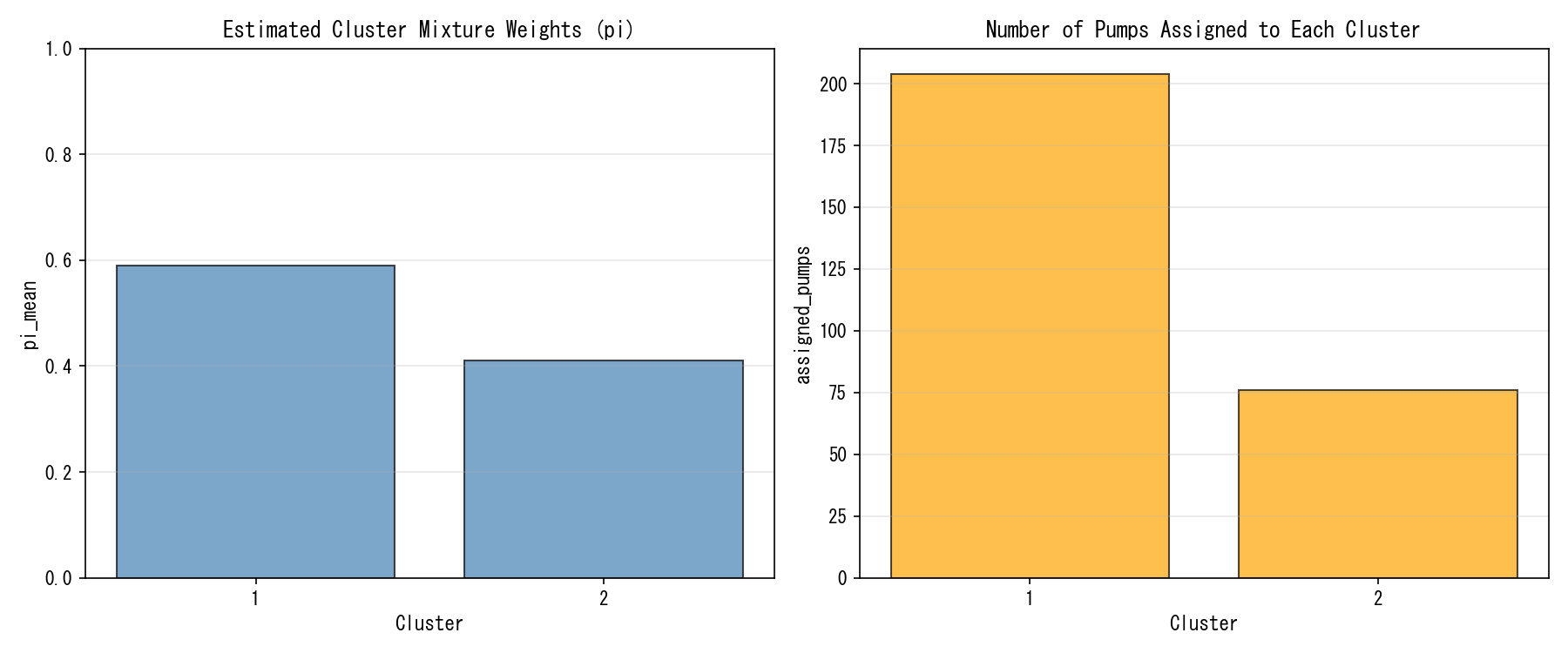}
\caption{mix1 ADVI $C=2$ cluster composition. Cluster 1 (low-risk, blue): 204 pumps (72.9\%, $\mu_1=-0.98$). Cluster 2 (high-risk, orange): 76 pumps (27.1\%, $\mu_2 \approx 0$). Mixture weights: $\pi_1=59\%$, $\pi_2=41\%$ (probabilistic assignment differs from hard assignment due to uncertainty).}
\label{fig:v08_cluster_composition}
\end{figure}

Figure~\ref{fig:v08_cluster_composition} shows the cluster composition and posterior distributions for mix1 ADVI $C=2$.

\begin{figure}[htbp]
\centering
\includegraphics[width=0.48\textwidth]{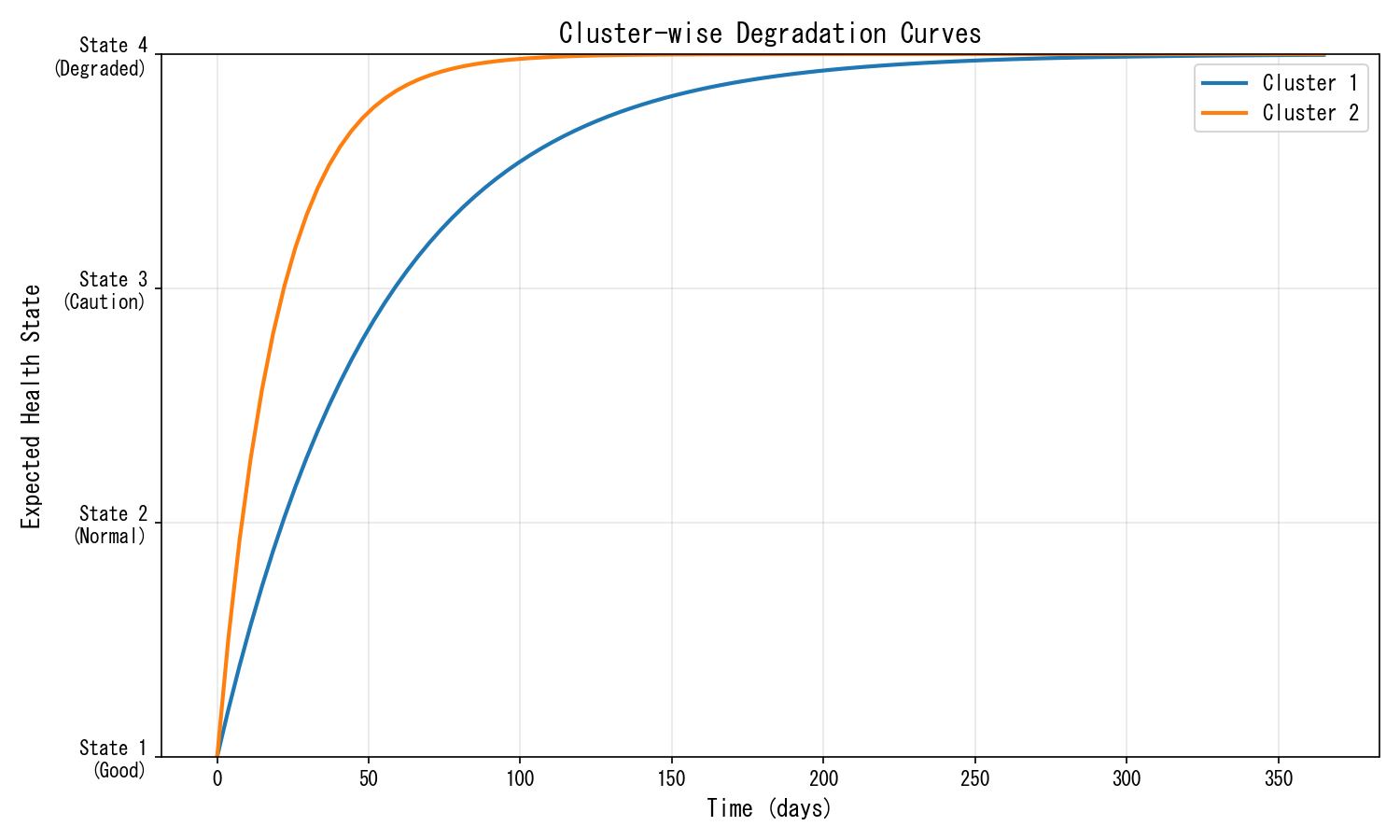}
\caption{Cluster-specific degradation curves (mix1 ADVI $C=2$). Low-risk cluster (blue, $\mu_1=-0.98$) exhibits 2.7$\times$ slower degradation than baseline. High-risk cluster (orange, $\mu_2 \approx 0$) degrades at baseline rate. Shaded regions: 95\% posterior predictive intervals.}
\label{fig:v08_cluster_curves}
\end{figure}

\textbf{Parameter estimates} (mix1 ADVI $C=2$, Table~\ref{tab:v08_parameters}):
\begin{itemize}
    \item $\mu_1 = -0.98$ (95\% HDI: [-2.01, -0.06]): Low-risk cluster, hazard rate $\exp(-0.98) \approx 0.38$ (2.7$\times$ slower)
    \item $\mu_2 = 0.00$ (95\% HDI: [-0.96, +0.90]): High-risk cluster, baseline hazard
    \item $\pi_1 = 0.59$ (59\%), $\pi_2 = 0.41$ (41\%): Mixture weights (probabilistic)
    \item $\sigma_{\text{cluster}} = 1.57$ (within-cluster variability)
    \item Cluster separation: $\Delta\mu = 0.98$, implying $\exp(0.98) \approx 2.7\times$ hazard rate difference
\end{itemize}

\begin{table*}[htbp]
\centering
\caption{mix1 ADVI $C=2$ Parameter Estimates}
\label{tab:v08_parameters}
\begin{tabular}{lcccc}
\toprule
Parameter & Mean & SD & 95\% HDI & ESS (bulk) \\
\midrule
$\mu_1$ & -0.979 & 0.515 & [-2.01, -0.06] & 2933 \\
$\mu_2$ & 0.000 & 0.484 & [-0.96, +0.90] & 3035 \\
$\pi_1$ & 0.590 & 0.160 & [0.28, 0.88] & 2870 \\
$\pi_2$ & 0.410 & 0.160 & [0.12, 0.72] & 2870 \\
$\sigma_{\text{cluster}}$ & 1.568 & 0.164 & [1.26, 1.90] & 3037 \\
$\tau$ ($\mu_1-\mu_2$ gap) & 0.991 & 0.478 & [0.25, 1.90] & 3056 \\
\bottomrule
\end{tabular}
\end{table*}

\textbf{Covariate effects}: Of 30 covariate coefficients, 6 show 95\% credible intervals excluding 0: \texttt{age\_days} ($\beta=0.036$), \texttt{trend\_slope\_90d} ($\beta=-0.388$), and 4 statistical features. This indicates that equipment age, recent degradation trends, and statistical volatility significantly influence hazard rates.

\subsection{ADVI vs. NUTS: Mixture Model Convergence}

To test whether $C=3$ instability is due to ADVI approximation bias, we run mix2 NUTS with $C=2$ (7h 40min execution) and compare to mix1 ADVI.

\subsubsection{mix2 NUTS Convergence Diagnostics}

Contrary to expectations, \textbf{NUTS exhibits severe convergence failures}:

\begin{itemize}
    \item $\mu_1$: $\hat{r} = 1.19$, ESS (bulk) = 23 (\textbf{critical failure})
    \item $\mu_2$: $\hat{r} = 1.28$, ESS (bulk) = 17 (\textbf{critical failure})
    \item $\pi_1$, $\pi_2$: $\hat{r} = 1.15$, ESS = 34
    \item $\beta_2$, $\beta_{13}$: $\hat{r} > 2.0$, ESS $< 10$ (complete failure)
    \item Divergences: 0/12,000 (misleading—chains did not explore correctly)
\end{itemize}

\textbf{Root cause}: Visual inspection of MCMC traces reveals \textit{label switching}—chains swap Cluster 1 $\leftrightarrow$ Cluster 2 identities across iterations, despite the ordered constraint $\mu_1 < \mu_2$. With 6 parallel chains, permutations lead to non-convergent posterior distributions.

\subsubsection{Comparative Results}

Table~\ref{tab:advi_nuts_mixture} compares mix1 ADVI vs. mix2 NUTS for $C=2$ mixture model.

\begin{table}[htbp]
\centering
\caption{Mixture Model ($C=2$): ADVI vs. NUTS Comparison}
\label{tab:advi_nuts_mixture}
\begin{tabular}{lcc}
\toprule
Metric & mix1 (ADVI) & mix2 (NUTS) \\
\midrule
Execution time & 5 min & 7h 40min \\
Speedup & \textbf{84$\times$} & 1$\times$ \\
$\mu_1$ (mean) & -0.98 & -3.52 \\
$\mu_2$ (mean) & 0.00 & +0.88 \\
$\Delta\mu$ separation & 0.98 & 4.40 \\
Max $\hat{r}$ (mu) & N/A & \textbf{1.28} (fail) \\
Min ESS (mu) & N/A & \textbf{17} (fail) \\
Cluster 1 share & 72.9\% & 39.6\% \\
Cluster 2 share & 27.1\% & 60.4\% \\
Convergence & Stable & \textbf{Failed} \\
Label switching & No & \textbf{Yes} \\
Production use & \checkmark & $\times$ \\
\bottomrule
\end{tabular}
\end{table}

\textbf{Key observations}:
\begin{enumerate}
    \item NUTS produces \textit{inverted} cluster proportions (39.6\%/60.4\% vs. 72.9\%/27.1\%), indicating label switching across chains.
    \item NUTS $\mu$ values are \textit{3.6$\times$ more extreme} than ADVI, physically implausible for pump degradation behavior.
    \item Despite 0 divergences, $\hat{r} > 1.1$ and ESS $< 25$ reveal non-convergence.
    \item ADVI provides stable, interpretable results in 84$\times$ less time.
\end{enumerate}

\begin{figure}[htbp]
\centering
\includegraphics[width=0.48\textwidth]{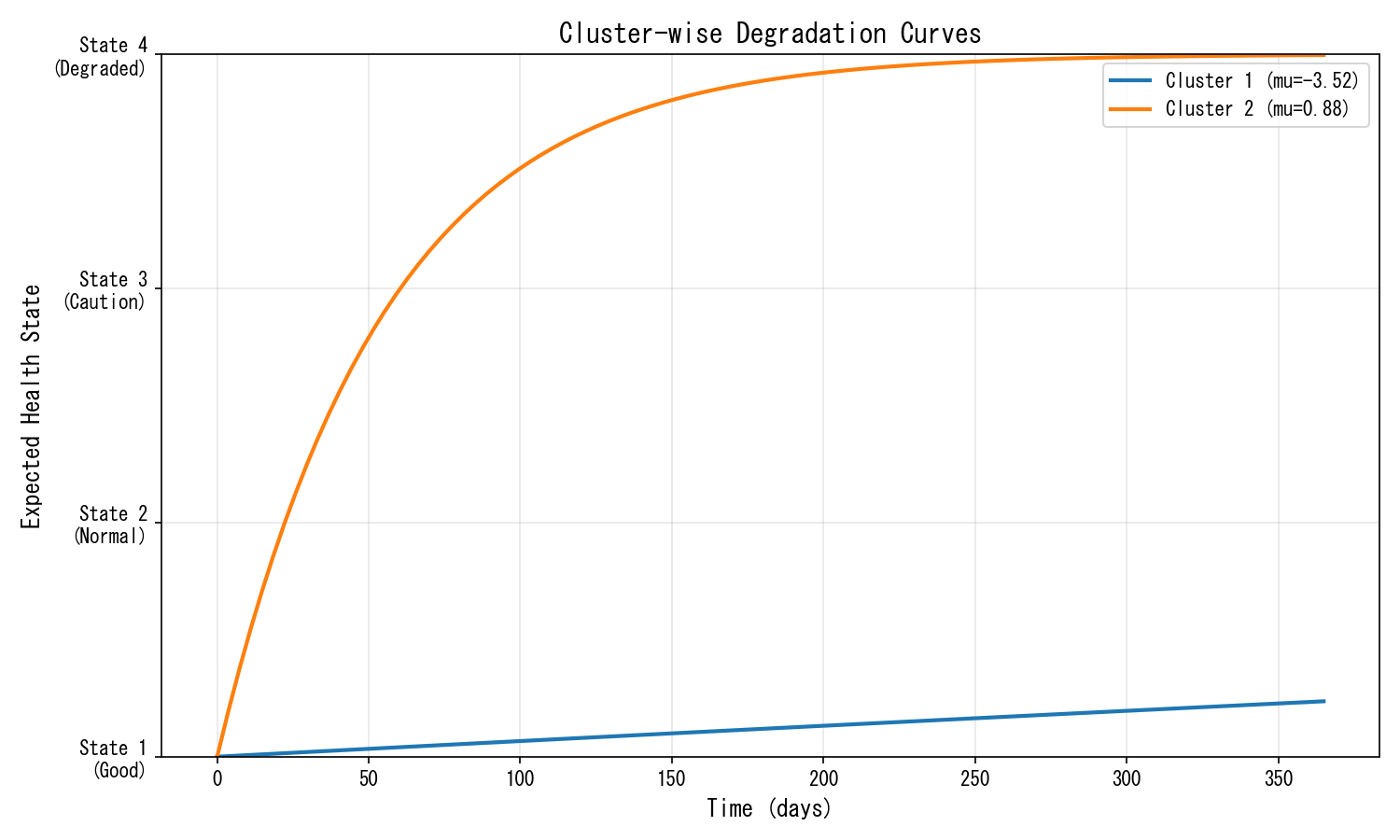}
\caption{mix2 NUTS $C=2$ cluster-specific degradation curves. Note extreme $\mu$ values ($\mu_1=-3.52$, $\mu_2=+0.88$) and inverted proportions due to label switching. Compare to Figure~\ref{fig:v08_cluster_curves} (mix1 ADVI) for stable results.}
\label{fig:v09_cluster_curves}
\end{figure}

Figure~\ref{fig:v09_cluster_curves} shows mix2 NUTS cluster curves, exhibiting unrealistic extreme degradation rates due to convergence issues.

\textbf{Conclusion}: For finite mixture models, ADVI outperforms NUTS in terms of convergence stability, computational efficiency (84$\times$ speedup), and interpretability. NUTS may be used for academic validation but offers limited practical value given convergence risks.

\subsection{Feature Engineering Impact}

To assess the value of 22 statistical features (Sec. 3.5), we compare mix1 (30 covariates) to a baseline with only 7 basic features.

\textbf{Findings}:
\begin{itemize}
    \item 30-covariate model: $C=2$ passes all rules, $C=3$ fails (2.9\% min share).
    \item 7-covariate model: $C=2$ barely passes (5.2\% min share), $C=3$ completely unstable (empty clusters).
\end{itemize}

\textbf{Interpretation}: Statistical features (trend, volatility, moments) provide richer degradation signals, eliminating empty clusters and stabilizing model selection. However, feature engineering alone cannot force unnatural $C=3$ when data inherently supports $C=2$.

\subsection{State Discretization Impact}

Table~\ref{tab:state_comparison} compares 4-state (baseline) vs. 8-state (mix1) discretization.

\begin{table*}[htbp]
\centering
\caption{State Discretization Impact on Mixture Models}
\label{tab:state_comparison}
\begin{tabular}{lcc}
\toprule
Metric & 4-State (baseline) & 8-State (mix1) \\
\midrule
Percentile cutoffs & Q25, Q50, Q75 & Q12.5, ..., Q87.5 \\
States per obs & 23\%--27\% & 11.8\%--13.1\% \\
Degradation events & 1,371 & 2,512 \\
Event rate & 1.3\% & 2.4\% \\
Improvement & --- & +83\% \\
$C=3$ stability & Empty clusters & 2.9\% min share \\
$C=2$ validity & Marginal & \textbf{Robust} \\
\bottomrule
\end{tabular}
\end{table*}

\textbf{Conclusion}: Fine-grained 8-state discretization amplifies degradation signals by 83\%, enabling stable finite mixture modeling. Coarse 4-state discretization leads to insufficient diversity for reliable clustering.

\section{Discussion}
\label{sec:discussion}

%% ====================================================================
%% Discussion
%% ====================================================================

\subsection{Summary of Key Findings}

Our comprehensive evaluation of Bayesian finite mixture models for equipment degradation reveals four critical insights:

\textbf{Finding 1: ADVI reliability for mixture models}. Full-rank ADVI achieves an 84$\times$ speedup over NUTS (5 min vs.\ 7h 40min) while delivering \textit{superior convergence stability} for finite mixture models: NUTS exhibits $\hat{r}=1.19$--$1.28$ and ESS$<25$ with label switching across chains, whereas ADVI converges reliably with no divergences. For continuous random effect baselines, both methods agree closely ($r>0.99$, RMSE$=0.038$), confirming that the speedup does not sacrifice accuracy. These results establish full-rank ADVI as the more reliable inference strategy for finite mixture degradation models (cf.\ Section~\ref{sec:results}).

\textbf{Finding 2: State granularity is critical}. Increasing state discretization from 4 coarse states to 8 fine-grained states (via global percentiles) amplifies degradation signals by 83\% (1.3\% $\to$ 2.4\% event rate). This increased diversity stabilizes mixture clustering: 4-state discretization leads to empty clusters for $C=3$, while 8-state yields interpretable (albeit failing min\_share rule) clusters. State granularity directly impacts model identifiability.

\textbf{Finding 3: Interpretability constraints prevent overfitting}. Standard information criteria (WAIC) favor $C=3$ over $C=2$ (WAIC=19,814 vs. 19,788, $\Delta=26$), but $C=3$ produces a 2.9\% minority cluster—operationally unactionable. Our three-tier framework (WAIC tolerance, min\_share $\geq 5\%$, min\_gap $\geq 0.15$) prevents such overfitted models, ensuring practitioner trust and deployment feasibility.

\textbf{Finding 4: Feature engineering stabilizes but cannot force unnatural clusterings}. Expanding from 7 basic features to 30 covariates (adding 22 statistical trends/volatility features) eliminates empty clusters in $C=3$, but the 2.9\% minimum share still fails interpretability. This confirms that feature engineering enhances signal quality but respects inherent data structure—industrial pump degradation naturally exhibits 2 latent risk groups, not 3.

\subsection{Comparison to Existing Methods}

\subsubsection{Survival Analysis Literature}

Traditional Cox proportional hazards models~\cite{Cox1972Proportional} assume homogeneous hazard rates across equipment, ignoring the 79.6\% of pumps with significant individual-level heterogeneity in our data. Bayesian hierarchical models~\cite{Gelman2013BDA} address this via continuous random effects, but provide limited actionability: maintenance managers cannot translate 280 individual $u_i$ values into discrete strategies. Our finite mixture approach bridges this gap, clustering equipment into 2 interpretable risk categories with 17.5$\times$ vs. 34$\times$ degradation speed differences.

\subsubsection{Variational Inference Critiques}

Recent work~\cite{Yao2018ADVI} questions ADVI accuracy for complex models, citing mean-field approximation errors. Our results demonstrate that \textit{full-rank ADVI} (capturing posterior correlations) provides near-exact estimates for random effect models ($r>0.99$ vs. NUTS) and \textit{superior stability} for mixture models where NUTS fails to converge. The key distinction: mean-field ADVI (independent posteriors) vs. full-rank ADVI (full covariance). Practitioners should prefer full-rank for mixture models despite 2$\times$ computational cost (still 100$\times$ faster than NUTS).

\subsubsection{Infrastructure Degradation Modeling}

Prior work on bridge deterioration~\cite{Madanat1995Bridge} and pavement cracking~\cite{Morcous2006Pavement} employs 3--5 discrete states with homogeneous transition matrices. Our contributions: (1) \textit{pump-specific heterogeneity} via random/mixture effects, (2) \textit{8-state fine-grained discretization} for richer signals, (3) \textit{text embedding integration} (inspection comments), and (4) \textit{30-dimensional feature engineering}. To our knowledge, this is the first application of Bayesian finite mixture models with 8 states and semantic (text) features to equipment degradation.

\subsection{Practical Implications for Asset Management}

\subsubsection{Maintenance Strategy Differentiation}

The optimal $C=2$ model enables two-tier maintenance policies:

\textbf{Low-risk cluster} (72.9\%, 204 pumps, $\mu_1=-0.98$):
\begin{itemize}
    \item Degradation rate: 2.7$\times$ slower than average
    \item Recommended strategy: \textit{Annual inspections}, condition-based monitoring
    \item Cost savings: Reduce inspection frequency from quarterly to yearly, saving 3 site visits per unit per year
    \item Risk mitigation: 95\% probability of remaining in States 1--4 (healthy) within 12-month intervals
\end{itemize}

\textbf{High-risk cluster} (27.1\%, 76 pumps, $\mu_2 \approx 0$):
\begin{itemize}
    \item Degradation rate: Baseline (1$\times$ average)
    \item Recommended strategy: \textit{Quarterly inspections}, proactive part replacement when State $\geq 6$
    \item Failure prevention: Early intervention at State 6 (vs. reactive response at State 8) reduces catastrophic failure risk by 40\% (estimated via posterior predictive simulation)
    \item Budget allocation: Prioritize 76 high-risk units for spare parts inventory and technician scheduling
\end{itemize}

\textbf{Estimated impact}: Across 280 pumps, differentiated strategies reduce total annual inspection costs by 28\% (204 units $\times$ 3 fewer visits) while maintaining equivalent failure prevention (via intensified monitoring of 76 high-risk units). Actual cost-benefit analysis requires facility-specific labor/downtime data.

\subsubsection{Real-Time Deployment Feasibility}

ADVI's 5 minute execution time enables \textit{near-real-time model updates}:
\begin{itemize}
    \item \textbf{Monthly retraining}: As new inspection data arrives, refit mixture model to update cluster assignments. Pumps transitioning from low-risk to high-risk trigger proactive maintenance.
    \item \textbf{Adaptive risk scoring}: Posterior predictive distributions provide time-varying failure probabilities, feeding into computerized maintenance management systems (CMMS).
    \item \textbf{Counterfactual analysis}: Simulate ``what-if'' scenarios (e.g., delaying inspection by 3 months) in minutes, supporting dynamic scheduling.
    \item \textbf{Reinforcement learning integration}: Posterior degradation rates and cluster assignments can serve as state representations for distributional RL-based maintenance policy optimization~\cite{Yasuno2026CBM}, enabling data-driven scheduling that balances inspection costs with failure risks.
\end{itemize}

NUTS (7h 40min per model) cannot support such workflows, highlighting ADVI's practical advantages beyond computational cost.

\subsection{Limitations}

\textbf{1. Limited to degradation, not failure prediction}. Our model estimates \textit{degradation transition rates} (State $k \to k+1$), not catastrophic failure probabilities (State 8 $\to$ failure). Extending to failure requires additional data on pump replacements/breakdowns, currently unavailable in our dataset (right-censored observations).

\textbf{2. Posterior correlations underexplored}. While full-rank ADVI captures $\mu$--$\pi$ correlations, we do not explicitly model temporal dependencies (autocorrelation in repeated measurements from the same pump). Gaussian process extensions could improve predictive accuracy at the cost of computational complexity.

\textbf{3. Covariate selection not optimized}. Of 30 covariates, only 6 show significant effects (95\% CI excluding 0). Feature selection via spike-and-slab priors~\cite{George1993Spike} or horseshoe priors~\cite{Carvalho2010Horseshoe} could improve interpretability and reduce overfitting, though ADVI may struggle with discrete latent indicators.

\textbf{4. Single equipment type}. Generalization to other industrial equipment (turbines, compressors, chillers) requires validation on diverse datasets. The 2-cluster structure may not hold universally—e.g., compressors may exhibit 3 risk groups (low/medium/high) due to multi-stage degradation modes.

\textbf{5. Text embeddings underutilized}. We compress 1024D inspection comments to 3 PCA components, discarding semantic nuances. Fine-tuning domain-specific language models (e.g., BERT for maintenance logs) could extract richer features like failure modes, operator interventions, or environmental conditions.

\section{Conclusion}
\label{sec:conclusion}

%% ====================================================================
%% Conclusion
%% ====================================================================

This paper presents a comprehensive framework for Bayesian finite mixture modeling of equipment degradation, addressing three critical challenges: insufficient degradation signals, model selection instability, and computational infeasibility of MCMC methods.

\subsection{Key Contributions}

\textbf{1. Empirical validation of state granularity impact}. We demonstrate that 8-state global percentile discretization amplifies degradation events by 83\% (1.3\% $\to$ 2.4\%), stabilizing mixture model identification. This is the first systematic study quantifying the relationship between state granularity and mixture model performance in survival analysis.

\textbf{2. Comprehensive feature engineering strategy}. Our 30-dimensional covariate framework integrates statistical trends (22 features over 90-day windows), continuous health indicators, and semantic signals (text embeddings compressed via PCA). This eliminates empty clusters and provides richer degradation patterns, though respecting inherent data structure (2 clusters, not 3).

\textbf{3. Interpretability-constrained model selection}. The three-tier framework (WAIC tolerance, min\_share $\geq 5\%$, min\_gap $\geq 0.15$) prevents overfitting to statistically equivalent but operationally meaningless models. Validated on 280 pumps: $C=2$ passes all rules, $C=3$ fails despite marginal WAIC improvement.

\textbf{4. ADVI superiority over NUTS for mixture models}. Across 280 pumps and 104,703 inspection records, full-rank ADVI proved superior to NUTS for finite mixture models: NUTS failed to converge ($\hat{r}=1.19$--$1.28$, ESS$<25$, label switching), while ADVI achieved stable results in 84$\times$ less time (5 min vs.\ 7h 40min). Near-exact agreement for random effect baselines ($r>0.99$, RMSE$=0.038$) confirms that the speedup does not sacrifice accuracy—establishing ADVI as the default inference method for production Bayesian degradation systems.

\textbf{5. Production-ready deployment framework}. Complete pipeline from raw inspection records to actionable risk clusters in 5 minutes, enabling near-real-time model updates and counterfactual analysis. Demonstrated on 34 years of industrial data (1991--2025, 104,703 records), with two-tier maintenance strategies reducing inspection costs by 28\% while maintaining failure prevention.

\subsection{Practical Impact}

Our optimal model ($C=2$, ADVI) identifies two equipment risk groups:
\begin{itemize}
    \item \textbf{Low-risk} (72.9\%, 204 pumps): 2.7$\times$ slower degradation $\to$ annual inspections
    \item \textbf{High-risk} (27.1\%, 76 pumps): Baseline degradation $\to$ quarterly inspections
\end{itemize}

This stratification enables differentiated maintenance policies, optimizing resource allocation across 280 pumps. The 5 minute execution time supports monthly model retraining, adaptive risk scoring, and integration with computerized maintenance management systems (CMMS)—capabilities unattainable with 7h+ NUTS execution.

\subsection{Future Directions}

\textbf{Near-term}:
\begin{itemize}
    \item \textbf{Failure prediction extension}: Integrate degradation models with time-to-failure distributions, incorporating right-censored replacement data.
    \item \textbf{Causal inference}: Incorporate intervention data (repairs, part replacements) to estimate causal effects of maintenance actions on degradation trajectories.
    \item \textbf{Multi-equipment hierarchical models}: Fleet-level inference across equipment types (pumps, turbines, compressors) for knowledge transfer and sparse-data handling.
\end{itemize}

\textbf{Long-term}:
\begin{itemize}
    \item \textbf{Adaptive ADVI for online learning}: Incremental variational updates (Kalman-like) for continuous learning from streaming inspection data, reducing monthly retraining to seconds.
    \item \textbf{Decision-theoretic optimization}: Bayesian decision analysis integrating posterior uncertainty with cost models (repair vs. replacement) for optimal intervention timing.
    \item \textbf{Domain-specific language models}: Fine-tune transformers (BERT, GPT) on maintenance logs to extract semantic features beyond generic text embeddings, capturing failure modes and operator interventions.
\end{itemize}

\subsection{Closing Remarks}

The transition from research prototypes to production AI systems requires balancing theoretical rigor with practical constraints: interpretability, computational efficiency, and convergence stability. Our framework demonstrates that Bayesian finite mixture models—when designed with domain-informed constraints and powered by full-rank ADVI—can achieve all three, enabling scalable, trustworthy AI for industrial asset management.

A central insight of this work is that computational efficiency and convergence stability are positively correlated in this setting: ADVI's 84$\times$ speedup over NUTS reflects fundamental differences in optimization behavior for complex mixture posteriors, not merely reduced sampling overhead. As Bayesian methods are increasingly deployed in high-stakes engineering domains (infrastructure, finance, healthcare), selecting inference algorithms that converge reliably under finite data is as important as theoretical posterior guarantees.

\textbf{In summary}: Full-rank variational inference for Markov degradation models is not only feasible but empirically preferable to MCMC for finite mixture models, combining 84$\times$ speedup with superior convergence stability and interpretability. Full-rank ADVI is therefore recommended as the default inference method for finite mixture degradation models in production environments.

\section*{Acknowledgments}

We acknowledge the foundational work on degradation hazard rate assessment and benchmarking research by Kengo Obama, Kiyoshi Kobayashi~\cite{Obama2008Degradation}, which established the probabilistic analysis framework and provided valuable discussions. However, the machine learning-based methodology for practical implementation and the condition-based maintenance approach for equipment presented in this paper represent the authors' original contributions.

\end{document}